\documentclass{article}

\usepackage{arxiv}

\usepackage[utf8]{inputenc} % allow utf-8 input
\usepackage[T1]{fontenc}    % use 8-bit T1 fonts
\usepackage{hyperref}       % hyperlinks
\usepackage{url}            % simple URL typesetting
\usepackage{booktabs}       % professional-quality tables
\usepackage{amsfonts}       % blackboard math symbols
\usepackage{nicefrac}       % compact symbols for 1/2, etc.
\usepackage{microtype}      % microtypography
\usepackage{lipsum}
\usepackage{amsmath}
\usepackage{graphicx}

\title{Service Selection using Predictive Models and Monte-Carlo Tree Search}

\author{
  Cliff Laschet\\
  Philips Research$^1$\\
  \texttt{clifflaschet@gmail.com} \\
   \And
  Jorn op den Buijs\\
  Collaborative Care Solutions\\
  Philips Research\\
  Eindhoven, The Netherlands \\
  \texttt{jorn.op.den.buijs@philips.com} \\
   \And
  Mark H. M. Winands\\
  Dept. of Data Science and Knowledge Engineering\\
  Maastricht University\\
  Maastricht, The Netherlands \\
  \texttt{m.winands@maastrichtuniversity.nl} \\
   \And
  Steffen Pauws \\
  Collaborative Care Solutions\\
  Philips Research\\
  Eindhoven, The Netherlands\\
  \texttt{steffen.pauws@philips.com}\\
  \\
  Dept. of Communication and Cognition\\
  Tilburg University\\
  Tilburg, The Netherlands\\
  \texttt{s.c.pauws@uvt.nl} \\
}

\begin{document}
\maketitle

\begin{abstract}
This article proposes a method for automated service selection to improve treatment efficacy and reduce re-hospitalization costs. A predictive model is developed using the National Home and Hospice Care Survey (NHHCS) dataset to quantify the effect of care services on the risk of re-hospitalization. By
taking the patient's characteristics and other selected services into account, the model is able to indicate the overall effectiveness of a combination of services for a specific NHHCS patient. The developed model is incorporated in Monte-Carlo Tree Search (MCTS) to determine optimal combinations of services that minimize the risk of emergency re-hospitalization. MCTS serves as a risk minimization algorithm in this case, using the predictive model for guidance during the search. Using this method on the NHHCS dataset, a significant reduction in risk of re-hospitalization is observed compared to the original
selections made by clinicians. An 11.89 percentage points risk reduction is achieved on average. Higher reductions of roughly 40 percentage points on average are observed for NHHCS patients in the highest risk categories. These results seem to indicate that there is enormous potential for improving service
selection in the near future.
\end{abstract}


\section{Introduction}
\footnotetext[1]{Cliff Laschet is now with Amazon, Germany}
\subsection{Problem}
The modern health care system is struggling to provide cost-effective and high-quality care to patients due to a growing number of treatment options and rapidly increasing costs. Even though there is a growing body of empirical data on treatment outcomes, it is rarely used effectively in a practical health care setting to tackle these challenges.\\

The empirical knowledge of when a patient should or should not receive a particular care service is not
readily available to the clinician when starting treatment. In most cases, these care services are chosen
based on personal experience, general consensus and subjective opinion. As a result, the clinician might
not take all relevant patient characteristics into account, or important interactions between characteristics and care services might be overlooked to determine treatment efficacy [1]. This can lead
to the formation of care plans that are not tailored to the individual, resulting in lacking treatment
efficacy and unnecessary costs. \\

In several health care settings, such as chronic disease management, treatment often consists of
providing a care plan to a patient. A care plan is essentially a combination of care services (e.g. physical therapy, nutrition advice and exercise) that aims to improve patient outcomes (e.g., survival rates, quality of life), while simultaneously reducing costs. For a care plan to have the best possible effect on a patient, it is extremely important to tailor the combination of care services to the patient’s
characteristics. \\

The process of composing such a care plan is also referred to as service selection. It is more challenging
compared to selecting individual care services, as combinations of services may have a larger
collaborative effect than the sum of its individual elements, suggesting that the interactions between
care services influence the overall treatment effectiveness as well. Hence, to estimate the overall
treatment effectiveness of a care plan, not only interactions between patient characteristics and care
services, but also interactions between multiple care services need to be taken into account. This is
illustrated in Figure 1, in which the interactions between care services and the interactions of a single
patient characteristic (age) with all care services are shown. In the current methodology, in which a
clinician estimates all of these interactions based on personal experience, creating highly effective care
plans by manual service selection is difficult if not impossible to achieve.

% FIGURE 1
\begin{figure}
  \centering
  \includegraphics[width=0.9\textwidth]{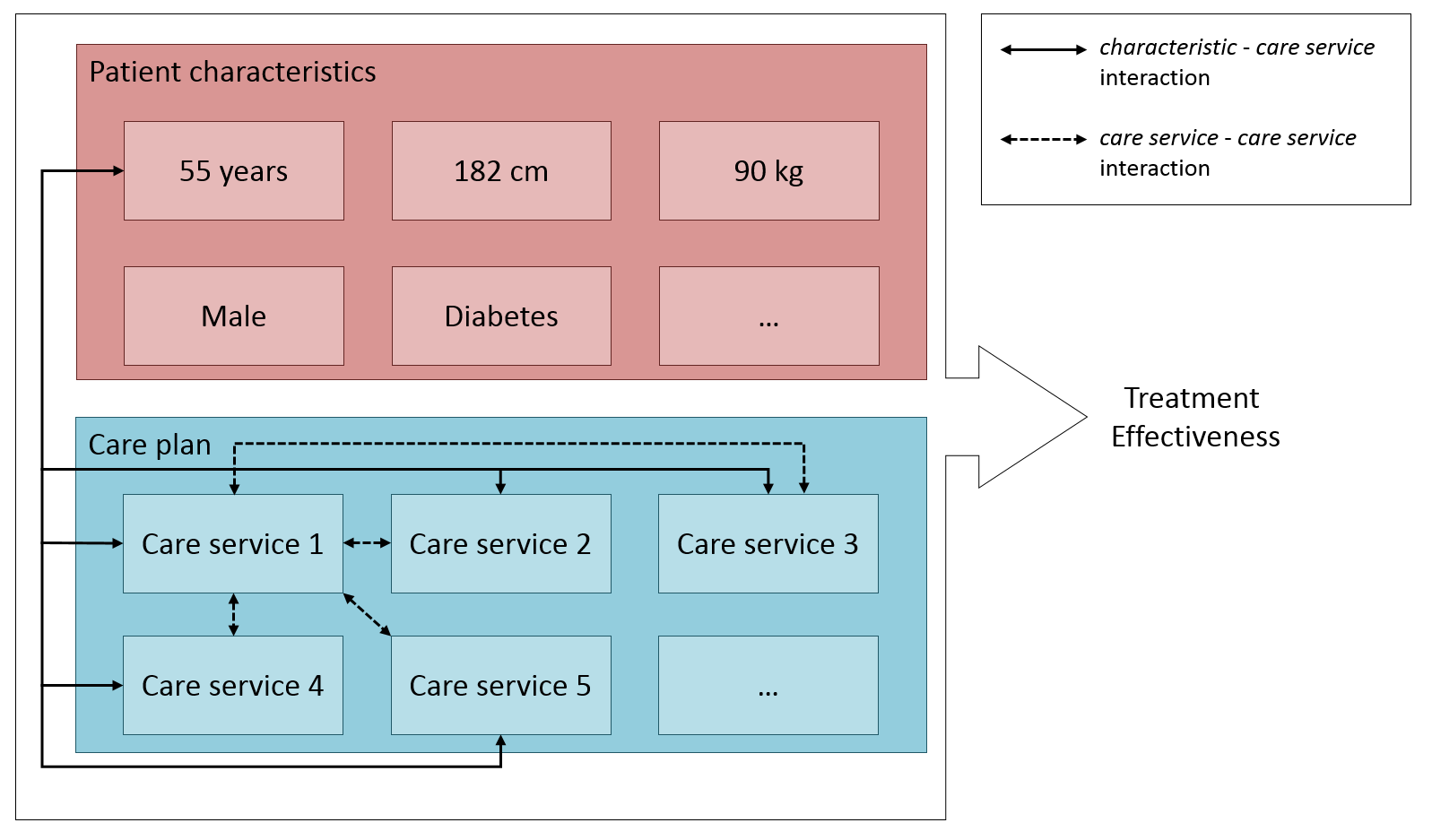}
  \caption{Interactions between care services may cause a care plan to have a larger collaborative effect than the sum of its individual components. Interactions between care services and patient characteristics influence the effectiveness of the care plan for a specific individual. Both sets of interactions may play an important role in creating highly effective care plans.}
  \label{fig:fig1}
\end{figure}

\subsection{Goal}
The increased usage of Electronic Medical Records (EMR) and availability of a variety of large public
(bio)medical datasets allow the application of novel techniques in the field of Artificial Intelligence (AI) to construct highly effective and tailored care plans for patients. These techniques take the individual’s characteristics and interactions with care services into account and may quantify the effects of these interactions on the treatment effectiveness. Hence, these techniques may be used for service selection. \\

This article proposes a method for improving service selection using predictive models and search. The
developed method is able to automatically recommend a care plan that significantly reduces the risk of
future undesirable health outcomes. The method presented herein is however not limited to health
outcomes specifically and may also be applied using different measures, such as treatment costs. In
order to create highly effective care plans, the method takes both patient characteristics and empirical
evidence of service efficacy into account while creating these combinations of care services.

\subsection{Related work}
The necessity and effectiveness of patient stratification for post-acute care has been discussed in the
work of Holland [2] and Bowles [1,3]. Holland investigated how to determine which patients in a hospital
would need a `comprehensive evaluation by a discharge planner'. If a patient is sent to a discharge
planner, a patient-specific service selection is manually performed by the discharge planner. This
research led to the development of a discharge tool that automatically stratified patients in groups
indicating whether or not they required an evaluation by the discharge planner. Validation of this tool in
several hospitals showed that, on average, the length of stay of patients in hospitals was reduced by
20\%. The work of Bowles continued on these results by investigating the development of a decision
support system. This system helps the discharge planner decide which of the referred patients should be
given post-acute care in order to prevent hospital readmissions. Validation of the decision support
system showed that the number of hospital readmissions dropped by 26\%. This clearly indicates the
potential impact of tailored post-acute care in healthcare.\\

The work of Bennett and Hauser [4] discusses the development of a general-purpose clinical artificial
intelligence framework for clinical decision making over timespans. It addresses similar issues in modern
healthcare and proposes an environment to simulate the effect of various treatment policies and
payment methodologies over time. The work presented in this article has similar motivations for
development and therefore also focuses on tailored medical treatment. The difference is however
situated in the problem specification: while the work of Bennett and Hauser focuses on a binary yes or
no treatment decision over time, this work focuses on selecting the best combination of care services at
the current moment in time.\\

Research in the implementation of automated service selection has been limited. As far as is known by
the authors, no research has been performed related to service selection that even generally indicates
how this process should be realized. In this sense, the work discussed in this article is novel and
exploratory. Moreover, the simulation based technique Monte-Carlo Tree Search has not yet been
combined with predictive models based on medical data.

\subsection{Current work}
This article tackles the service selection challenge by investigating the specific case of automated postacute service selection, in which the outcome of interest is the prevention of emergency rehospitalizations. After a patient is discharged from a hospital, several post-acute services are provided to lower the risk of re-hospitalization within a time frame of 60 days. Re-hospitalization is often far more expensive than the provision of these post-acute services [5]. By taking patient characteristics into
account, such as demographics, payment information and ICD9 codes, a care plan can be tailored to a
specific patient to lower the risk of re-hospitalization as much as possible. Currently, care plans are
composed by a nurse or physician shortly after the hospital discharge. When an emergency hospital
readmission does occur, it is a sign that the provided care plan was not able to sufficiently stabilize the
patient. The question to be asked in that case is whether a different combination of care services could
have prevented the re-admission, and more specifically, which combination this should have been. \\

In order to quantify the effectiveness of a care plan on a patient, a predictive model is developed from
clinical data describing service usage and outcome of a population of real patients, as shown in the left
half of Figure 2. The predictive model is a concrete representation of the interactions between multiple
care services and between care services and patient characteristics. Furthermore, it provides the
functionality of assigning a value to any care plan indicating the effectiveness of the care plan for a
specific patient and outcome of interest. The necessity of tailoring to the patient and the selection bias
of the clinical data are matters that need to be considered as well. This is discussed in more detail in
Subsections 2.1 and 2.2. \\

While the predictive model can evaluate the effectiveness of a given care plan, it cannot directly indicate
which care plan(s) would provide a good or even optimal outcome. Hence, the question remains which
services should be selected to form a care plan that receives the best evaluation by the predictive
model. The large number of possible services to choose from, plus the fact that these services have to
be combined, make this a challenging single-agent search problem. In the context of a search, the
problem of service selection can be described as a tree, where nodes represent care services and a path
from the root node through the tree represents a care plan. The predictive model is used as a utility
function during the search to indicate the quality of care plans given patient characteristics. This is
shown in Figure 2 and discussed in more detail in Subsection 2.3.

% FIGURE 2
\begin{figure}
  \centering
  \includegraphics[width=0.9\textwidth]{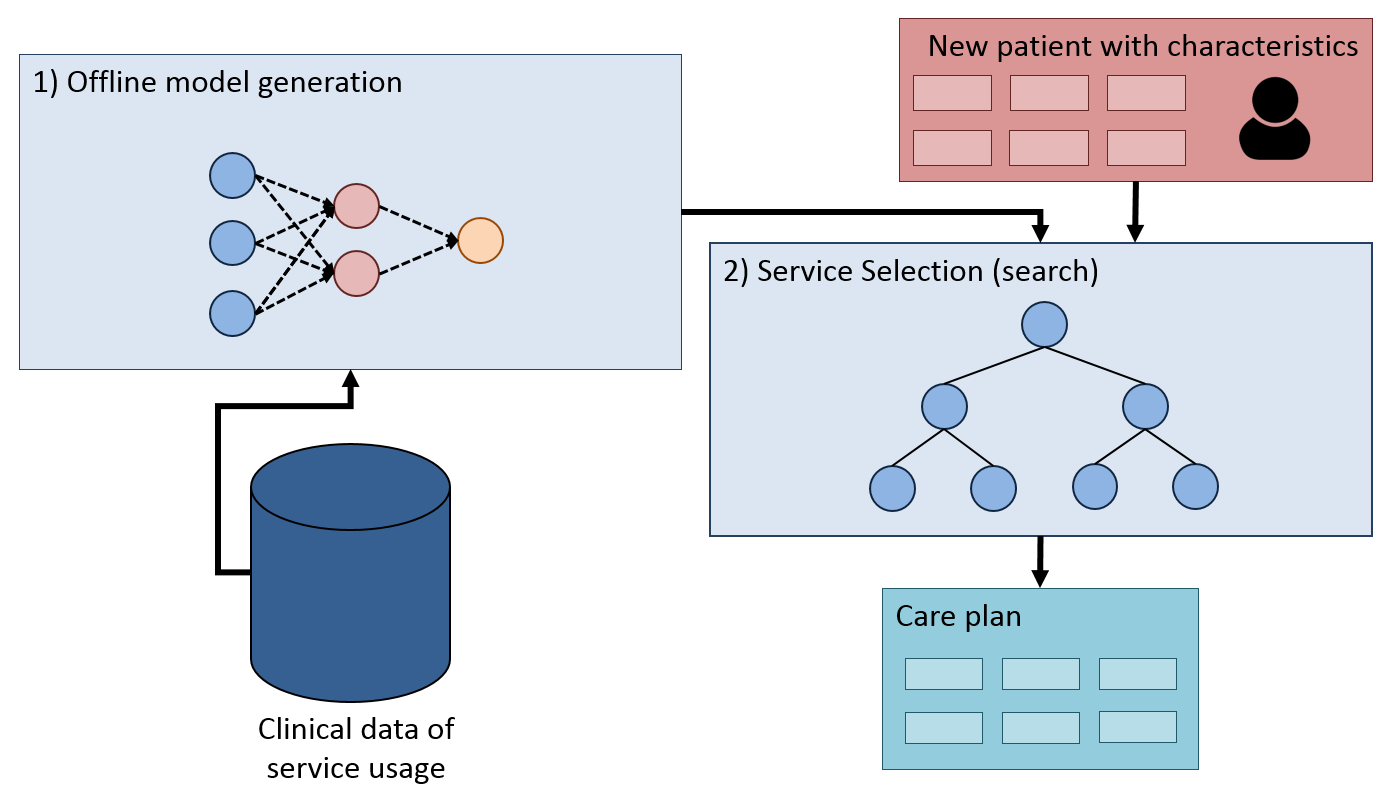}
  \caption{High-level overview of the automated service selection process. A predictive model is generated offline based on clinical data to predict care plan effectiveness, given a care plan and patient characteristics. In a separate procedure, the model is used as a utility function during search for optimal care plans, given a patient with several characteristics.}
  \label{fig:fig2}
\end{figure}

\section{Methods}
\subsection{Data}
The clinical data used for automated service selection was collected by the National Home and Hospice
Care Survey (NHHCS) in 2007 by the Centers for Disease Control and Prevention (CDC) group, located in
Atlanta, United States of America [6]. Data collection for NHHCS was approved by the NCHS Research
Ethics Review Board. The analysis of de-identified data from the survey is exempt from USA federal
regulations for the protection of human research participants. Furthermore, the analysis of restricted
data through the NCHS Research Data Center is also approved by the NCHS Ethics Review Board.\\

The data is a nationally representative sample of U.S. home health and hospice agencies, providing
information regarding their population of patients, staff and services. The data contained both home
care and hospice patients. Only home care patients were considered, as hospice patients are receiving
end-of-life care and are unlikely to recover. Home care patients are more likely to stabilize in health
after hospital discharge. Each home care patient received one or more post-acute care services after
being discharged from a hospital admission. \\

The dataset consists of 4683 home care patients. Each patient is described by roughly 300 features,
including an extensive set of demographics and financial information (e.g. insurance coverage and total
sum billed for treatment). A full overview of all the included features can be found on the website of the
CDC group [6]. Several studies already provided a thorough analysis on the demographics of the
population [7–9]. \\

Additionally, the dataset contains diagnoses of patients as determined by medical staff and the
procedures performed on patients (e.g. diabetes and hip replacement, respectively). Both are indicated
in the data by using ICD9 codes. The ICD9 codes in the NHHCS dataset may provide insight into the
relations between medical conditions and the effect of post-acute care services, which is relevant
information for service selection. In order to cope with the high dimensionality of the ICD9 data, only
the most frequently occurring ICD9 codes were used for analysis. \\

% FIGURE 3
\begin{figure}
  \centering
  \includegraphics[width=0.9\textwidth]{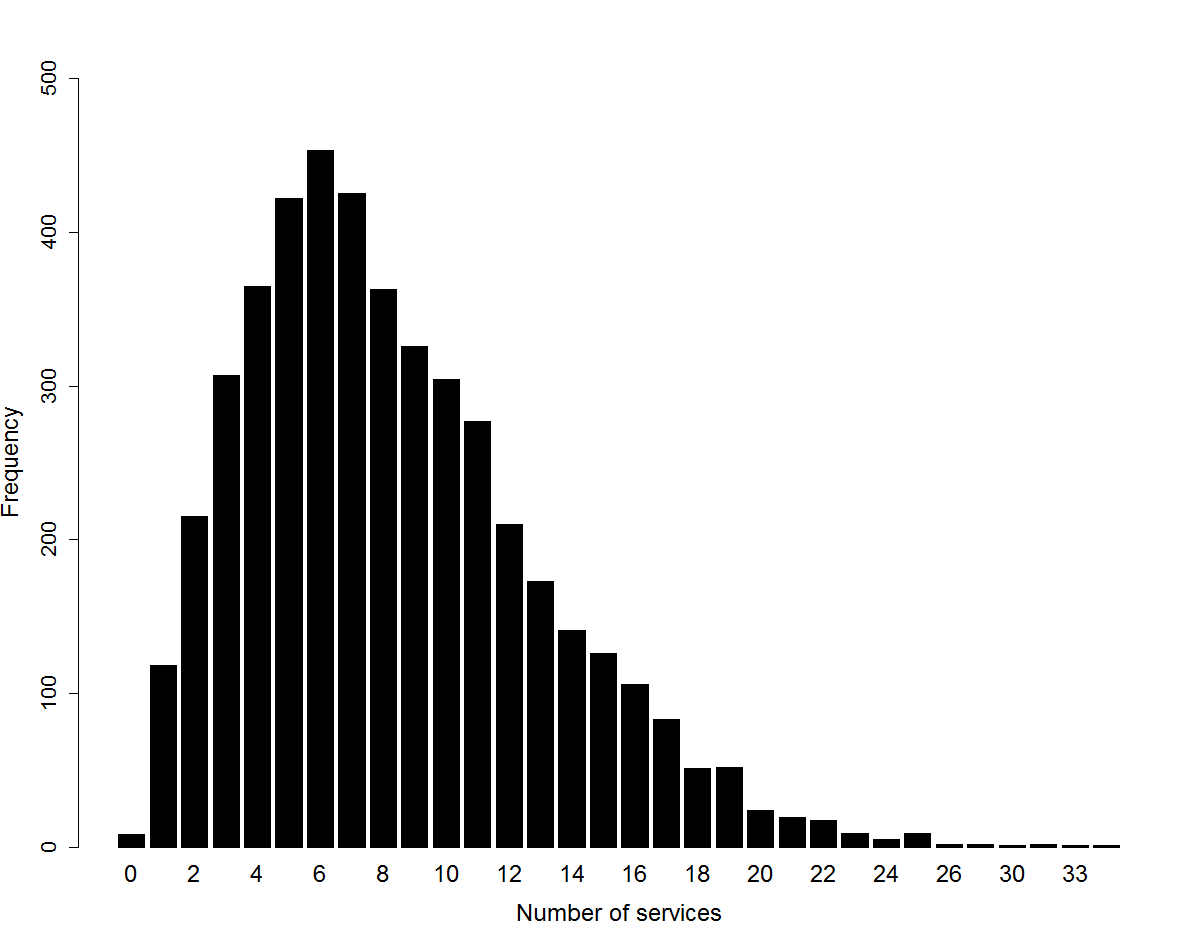}
  \caption{Distribution of the number of post-acute care services provided to home care patients.}
  \label{fig:fig3}
\end{figure}

The dataset also indicates the care plan provided to each patient after hospital discharge. The care
services are categorized into several groups, of which some examples are shown for each category in
Table 1. The dataset contains 69 distinct care services in total. As can be seen in Figure 3, most of the
patients are provided with 6 care services. One of the patients receives 34 care services, over half of the
available options. Eight patients receive no care services at all. The number of care services varies widely among the patient population. On average, a patient is provided with a combination of 8 care services. The most frequent care services are displayed in Table 2.\\

The goal of service selection is to select a care plan such that the care services as a combination have a
significant beneficial effect on the outcome of interest. Using the NHHCS dataset, the selected outcome
of interest is the need for emergent care (i.e. urgent and unplanned medical care) within the 60 days
following the start of receiving the selected care plan. \\

Because the NHHCS dataset is observational data, there is a selection bias regarding the use of care
services in the patient characteristics. For example, when comparing the group of patients that used a
walk cane with the group that did not, the distributions of patient characteristics will not resemble each
other. On a larger scale, patients that are more unstable health wise tend to use more care services
compared to more stable patients. The bias was partially removed by using propensity score matching
[10]. The propensity scores are used as weights to resample the data when fitting predictive models on
this data. By doing so, the data more closely resembles data generated from a randomized controlled
trial. \\

% TABLE 1
\begin{table}[]
\begin{center}
\caption{Examples of care services provided to home care patients.}
\label{tab:my-table1}
\begin{tabular}{|l|l|}
\hline
\textbf{Category} & \textbf{Examples} \\ \hline
Assistive & Walk cane, motorized cart, bed communication, shower grab bars. \\ \hline
Medical & IV infusion pump, oxygen, apnea monitor, glucose monitor. \\ \hline
Agency provided & Training and explanation of device usage. \\ \hline
Personal care & Transport, meals on wheels, volunteers. \\ \hline
Therapy & Physical therapy, speech therapy, occupational therapy. \\ \hline
Counselling & Dietary counselling, ethical issues counselling, spiritual services. \\ \hline
Services provided to family & Bereavement services, medication management. \\ \hline
\end{tabular}
\end{center}
\end{table}

% TABLE 2
\begin{table}[]
\begin{center}
\caption{Most frequent care services in the NHHCS dataset.}
\label{tab:my-table2}
\begin{tabular}{|l|l|}
\hline
\textbf{Service} & \textbf{Frequency} \\ \hline
Skilled services or nursing services & 3955 \\ \hline
Walker, cane or crutch & 2839 \\ \hline
Safety training services & 2343 \\ \hline
Medication management & 2205 \\ \hline
Agency supports with instruction, maintenance and monitoring of provided services & 1739 \\ \hline
Assistance with activity of daily living & 1679 \\ \hline
Shower chair or bath bench & 1653 \\ \hline
Activity of daily living & 1549 \\ \hline
Physical therapy & 1484 \\ \hline
Equipment instruction & 1387 \\ \hline
\end{tabular}
\end{center}
\end{table}

\subsection{Predictive models}
To find the care plan with the largest beneficial effect, the effect of a care plan on the outcome
of interest has to be quantified for a patient’s characteristics. Whether a care service should be provided
depends not only on the patient, but also on other care services that are provided by the plan as well. A combination of two or more care services may complement their individual effects, deteriorate their
effects, or make one or more care services in the plan redundant. Therefore, not only interactions
between patient characteristics and care services need to be taken into account, but also the
interactions between multiple care services. It is however challenging to quantify the effect all these
interactions have on the outcome. \\

The subfield of machine learning provides a way to quantify these effects. A predictive model is created
using the NHHCS survey results as training data. Because this data describes historic patient
characteristics, the care plan and outcome, the effects of patient characteristics and care service usage
on the outcome can be captured. The predictive model can then be used for the outcome prediction of
new care plans and patients that have not been encountered before. The remainder of this section first
discusses the requirements for creating a model that predicts emergent care given patient
characteristics and a care plan in Subsection 2.2.1. Next, the process of creating such a model is
discussed in Subsection 2.2.2.

\subsubsection{Model requirements}
Logistic regression [11] was chosen as a modeling technique, as it is suitable for the prediction of binary
outcomes following a binomial distribution, which is the case for the outcome of emergent care. The
predictors of the model are the interactions between patient characteristics and care services or
between two care services. Hence, the logistic regression model predicts the risk of receiving emergent
care, given the patient’s characteristics and a chosen care plan. \\

The characteristics of the patient (e.g. age) do not change during service selection and will therefore
have no influence by themselves on the result of creating a care plan. Thus, individual patient
characteristics are not considered as predictors for the model. Interactions between a patient
characteristic and a single care service (e.g. age with walking cane), or interactions between two care
services (e.g. walking cane with shower grab bars), are considered. By including patient characteristics in
these interactions, recommendations can be tailored to the patient. By including interactions between
two care services, the combinatory effect of care services is taken into account. Interactions of higher
degrees are not considered, due to limitations of the available variety in data. These are however partly
expressed by including all of the lower degree variations (e.g., \emph{x:y:z} is not included, but \emph{x:y}, \emph{y:z} and \emph{x:z} may be included).

\subsubsection{Ensemble logistic regression models}
When fitting the coefficients of the predictors in the model using data, a variation in values among the patient records is needed in order to quantify the effect of predictors on the outcome of the patient. Because the data used for generating the predictive model contains roughly 5000 patient records and 300 features, the limited number of records cannot express all the possible combinations of feature values. Hence, it is highly unlikely all combinations of features can be quantified by the available data. Moreover, interactions between services and patient characteristics may cause separations of the data that may result in severe overfitting of the model, inaccurate estimations of the predictor coefficients, or a failure to fit the model altogether. Finally, some interactions might have a strong predictive effect regarding the outcome, while others may not. This needs to be taken into account for feature selection. \\

A greedy approach similar to sequential backwards feature selection is used [12] with the following major additions:

\begin{itemize}
\item Several candidate feature sets are formed in parallel, where each process represents a group of selected features. 
\item A random shuffle of the remaining features is performed between iterations, moving features from one group to another.
\item Instead of using a single objective function for determining the quality of the selected features, each feature (i.e. predictor) is tested for significance as part of the set of predictors that form the model.
\end{itemize}

% FIGURE 4
\begin{figure}
  \centering
  \includegraphics[width=0.9\textwidth]{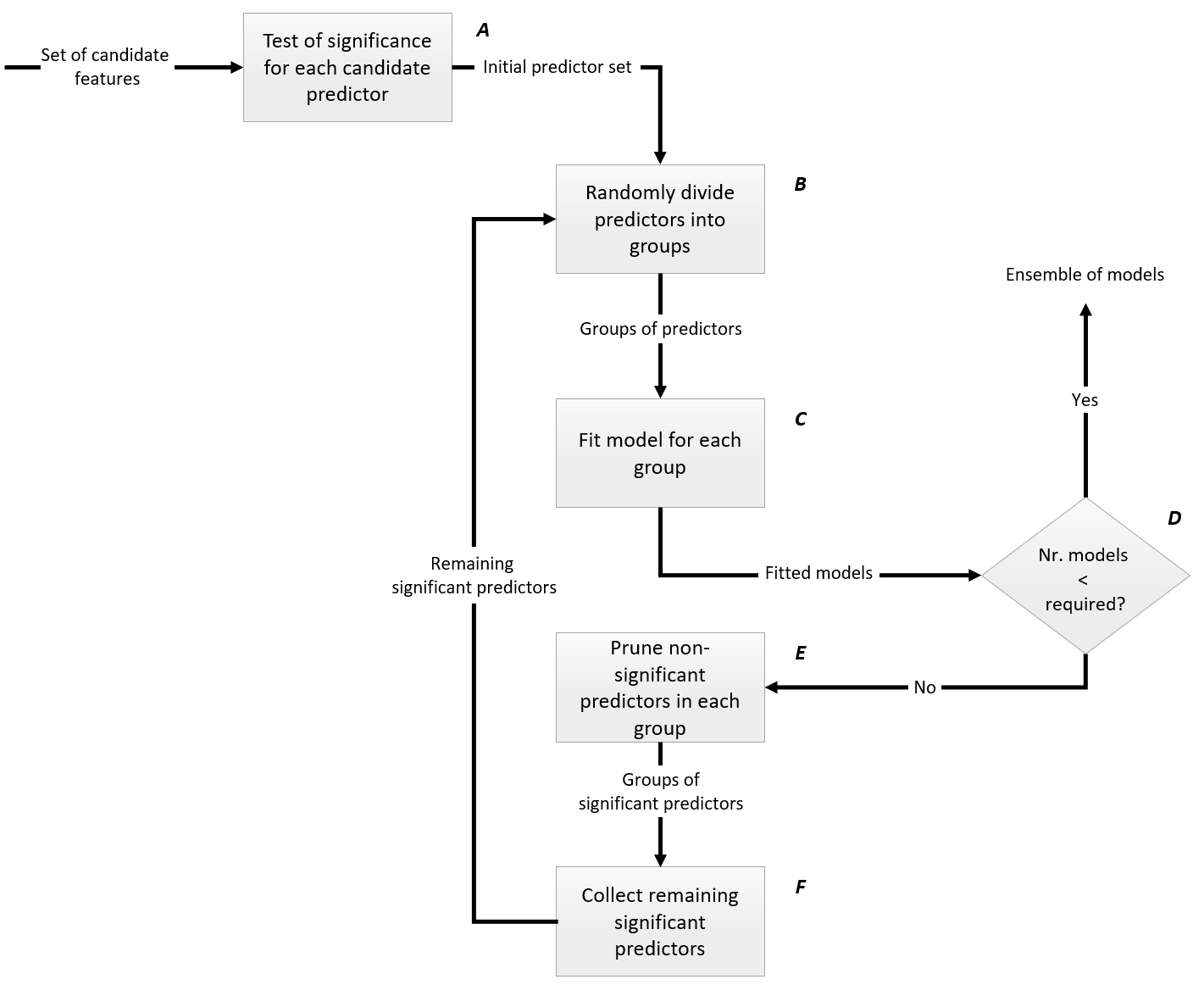}
  \caption{The feature selection process.}
  \label{fig:fig4}
\end{figure}

The structure of this process is shown in Figure 4. The process starts (A) by generating a set of individually significant interactions of features (e.g. \emph{age:cane}), using tests of significance (\(\alpha=0.05\)), regarding the outcome variable of emergent care. Any features directly related to the predicted outcome are not included. There are 272 individual features in the NHHCS dataset, not including any features related to the outcome of emergent care, of which 69 features represent the usage of a care service. This corresponds to roughly 19000 interactions between a care service and another feature that need to be tested for predictive power. Any predictor (i.e. interaction) for which no coefficient can be computed (due to a lack of variance in data) is discarded. This results in a set of 5449 distinct significant predictors, of which each describes either an interaction of two care services or one care service and a patient characteristic.\\

The feature selection continues (B) by randomly dividing the set of significant predictors in a predefined number of groups. Each group of predictors is used to generate a logistic regression model predicting emergency readmission (C). When combining these predictors into a single model, some of the predictors might become insignificant. Therefore, each model is pruned by removing any insignificant predictors (E). This results in a somewhat smaller group of predictors which still have predictive power when combined with other predictors. All remaining predictors are collected from each model, resulting in a single pruned set of predictors (F). This completes a single iteration.

\begin{subequations}
\begin{equation}
    n_i = \frac{|features|}{f_i} \\
\end{equation}
\begin{equation}    
    f_i = f_{i-1} + \frac{p_i}{n_{i-1}}
\end{equation}
\end{subequations}
 
Next, the process repeats by shuffling and regrouping the remaining predictors of each group into a smaller number of groups (B). By doing so, it is likely that over iterations a predictor will be grouped with a high variety of other predictors. The number of groups over iterations is determined as shown in Formula 1, where the number of groups \(n_i\) during iteration i is calculated by dividing the number of remaining features by the number of features per group \(f_i\) during iteration \(i\). The value of \(f_i\) is derived by summing the number of features per group of iteration \(i-1\) and the number of pruned features \(p_i\), divided by the number of groups during the previous iteration. The initial values for \(n_i\) and \(f_i\) are predefined. This process ensures that as long as predictors are being pruned, the number of groups will decrease over the iterations. By regrouping the remaining features into a smaller number of groups, models with an equal or larger number of predictors are generated. Over the course of these iterations, the algorithm slowly converges to fewer models with larger sets of predictors for each model, while also pruning any irrelevant predictors. The more predictors are pruned, the faster the algorithm converges to fewer, but larger models. This process is also illustrated in Figure 5. The feature selection stops when one model remains, or when a predefined number of models remain (D). Using the latter approach, the resulting set of models can be used as an ensemble model, where the score assigned by the ensemble model is the average over all scores returned by the individual models. \\

% FIGURE 5
\begin{figure}
  \centering
  \includegraphics[width=0.9\textwidth]{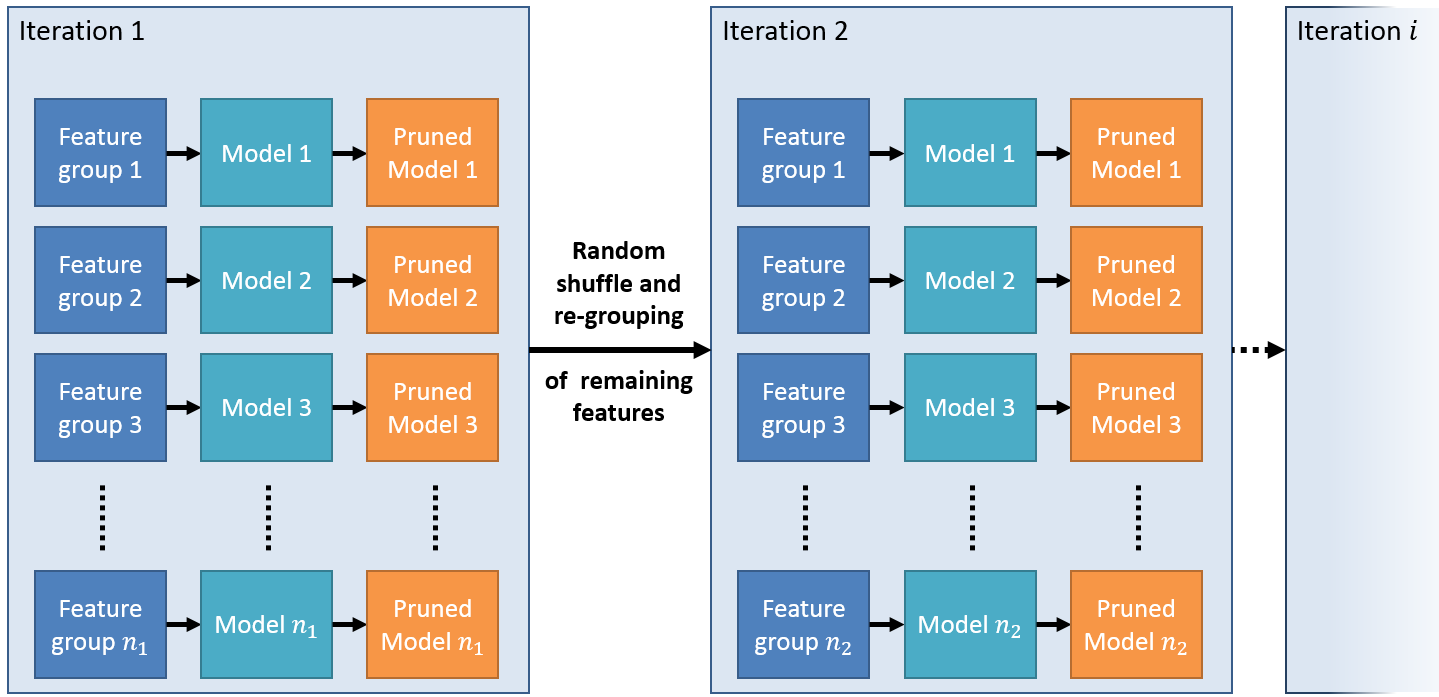}
  \caption{Generating an ensemble model by using a greedy approach, gradually converging from smaller groups of features to possibly larger pruned groups of features.}
  \label{fig:fig5}
\end{figure}

While the interactions between care services and patient characteristics are essential to the evaluation of a service combination, they do not provide any time-related information. The number of days that a home health patient has been receiving a certain care plan under home health care, also called length of stay (LOS), is possibly useful information for the efficacy of a care plan. Moreover, length of stay also indicates the stability of a patient in general. Therefore, the feature indicating the LOS was automatically included in every group during feature selection, forcing the inclusion of LOS as a predictor in the models. This resulted in slightly higher performance.

\subsection{Monte-Carlo Tree Search}
As was mentioned before, a patient receives a care plan consisting of 8 care services on average. Since there are 69 options, this problem is computationally expensive. Assuming 8 care services can be combined into a care plan, there are still more than \(8 \times 10^9\) distinct care plans. In a naive exhaustive search for all care plans, which allows for different orderings, there will be more than \(3.37 \times 10
^{14}\) leaf nodes. Preliminary tests showed that the time required for evaluating a care plan for a specific patient is relatively expensive (varying between 1 and 2 milliseconds), caused by the large number of predictors to be taken into account. Even when the evaluation of a single leaf node would take only 1 millisecond, a complete brute-force search would take more than 10,000 years to finish on a single-core machine, which is infeasible. Therefore, a heuristic search is required that quickly finds a care plan that reduces the risk of emergent care as much as possible, as soon as possible. \\

A popular technique for such single-agent search scenarios is  A* [13], a heuristic search technique that is best known for its widespread application in path finding and graph traversal, due to its high performance and accuracy. By performing a best-first search, it finds the least-cost path through a graph from a start node to a goal node, by following the path of the lowest expected cost, determined by a heuristic function. The heuristic function assigns admissible heuristic scores indicating the estimated cost from the current position to the goal node. In the case of service selection, the goal is not known. Therefore, it is hard to design an admissible heuristic function. This makes A* an impractical approach.

As it is difficult to construct a heuristic evaluator that can guide the search to promising care plans, a simulation based approach is chosen. The specific search technique used in this article is Monte-Carlo Tree Search (MCTS) [14,15], which has been successful in applications including optimization and scheduling problems, such as hospital planning [16] and planning of elective admissions for health care providers [17]. MCTS has also been a popular tree search algorithm in the research field of AI and games [18], most noticeably the game of Go [19]. It has outperformed other techniques such as genetic programming in many applications [20]. MCTS is a robust anytime search algorithm, which does not require an admissible evaluation function. MCTS does however require a utility function that can assign a quality score to any state. In the context of service selection, this utility function should be able to assess how a care plan would affect the risk of emergent care for a specific patient. To achieve this, the developed predictive model acts as a utility function during MCTS. By combining these two techniques, automated service selection based on empirical data can be implemented. \\

The remainder of this section is structured as follows. First, service selection is structured as a tree search in Subsection 2.3.1.  Next, MCTS is discussed in Subsection 2.3.2. Subsection 2.3.3 continues by showing how MCTS is implemented specifically for service selection. Finally, two MCTS enhancements are explained in Subsection 2.3.4.

\subsubsection{Problem representation}
Service selection can be structured as a single-agent search. The agent has a configuration, which is the currently selected care plan. To change the configuration, the agent can perform actions, which consists of adding care services to or removing care services from the care plan. This type of problem can be represented by a tree structure. A node in the search tree represents one of the 69 care services found in the NHHCS dataset. Actions are performed by transitioning from a tree node to its child nodes or parent node, which changes the care plan by adding or removing exactly one care service, respectively. Using this structure, a care plan is represented as a path through the tree, starting from the root node, which is illustrated by example in Figure 6. The root node represents an empty care plan. Children of a node are defined in such a way that repeats of already chosen care services are not allowed. Therefore, the branching factor of node \(n_d\) is \(69-d\), where \(d\) is the depth of tree in which the node is situated. However, the tree may still contain duplicate care plans, where the only difference is the ordering of the care services in the plan. It is assumed here that the order in which care services are chosen to form a care plan has no effect on the efficacy of the plan, as the care services will be provided simultaneously to the patient. If not mentioned otherwise, the size of the search tree is limited by allowing combinations of at most 8 services, as this is the average number of services patients received in the NHHCS dataset.

% FIGURE 6
\begin{figure}
  \centering
  \includegraphics[width=0.9\textwidth]{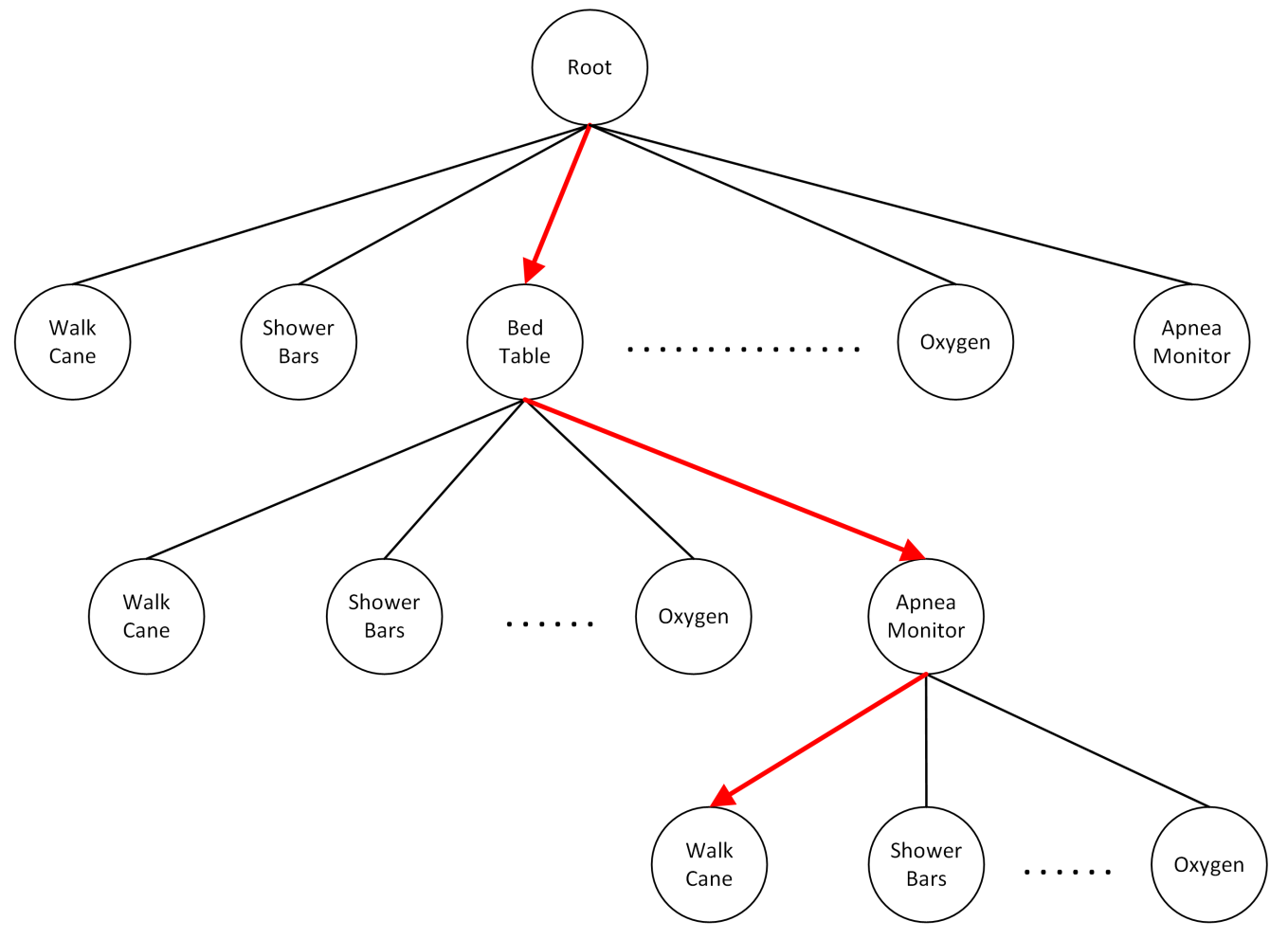}
  \caption{Representation of the service selection problem in a tree structure. Nodes represent individual care services. A path through the tree starting from the root node indicates a care plan.}
  \label{fig:fig6}
\end{figure}

\subsubsection{MCTS}
Monte-Carlo Tree Search (MCTS) [14,15] is a best-first search technique, which gradually builds up a search tree guided by Monte-Carlo simulations, either exploring undiscovered parts of the tree or exploiting already visited parts.  MCTS is a robust search algorithm; an admissible and accurate heuristic evaluation function is not required to find good solutions. Moreover, MCTS is anytime, which means it can be stopped at any given moment while still returning the best care plan found so far.\\

A schematic overview of MCTS is depicted in Figure 7. A simulation in MCTS can be divided in four phases [21]:
\begin{enumerate}
    \item In the selection phase states are encountered that are part of the search tree. Which states are encountered depends on the moves of the so-called selection strategy. Typically, the Upper Confidence bounds applied to Trees (UCT) is used [14,15].
    \item When a state is not in the tree, a simulation strategy chooses successor states until the end (i.e., the roll-out phase).
    \item MCTS then expands the tree by adding the first state it encountered along its roll-out (i.e., the expansion phase). The result of the simulation is then backpropagated to every node visited in the simulation up to the root node, updating node statistics accordingly for each node (i.e., the backpropagation phase). The node statistics are used by the selection strategy.
\end{enumerate}
These four phases are repeated until there is no time left for searching. 

% FIGURE 7
\begin{figure}
  \centering
  \includegraphics[width=0.9\textwidth]{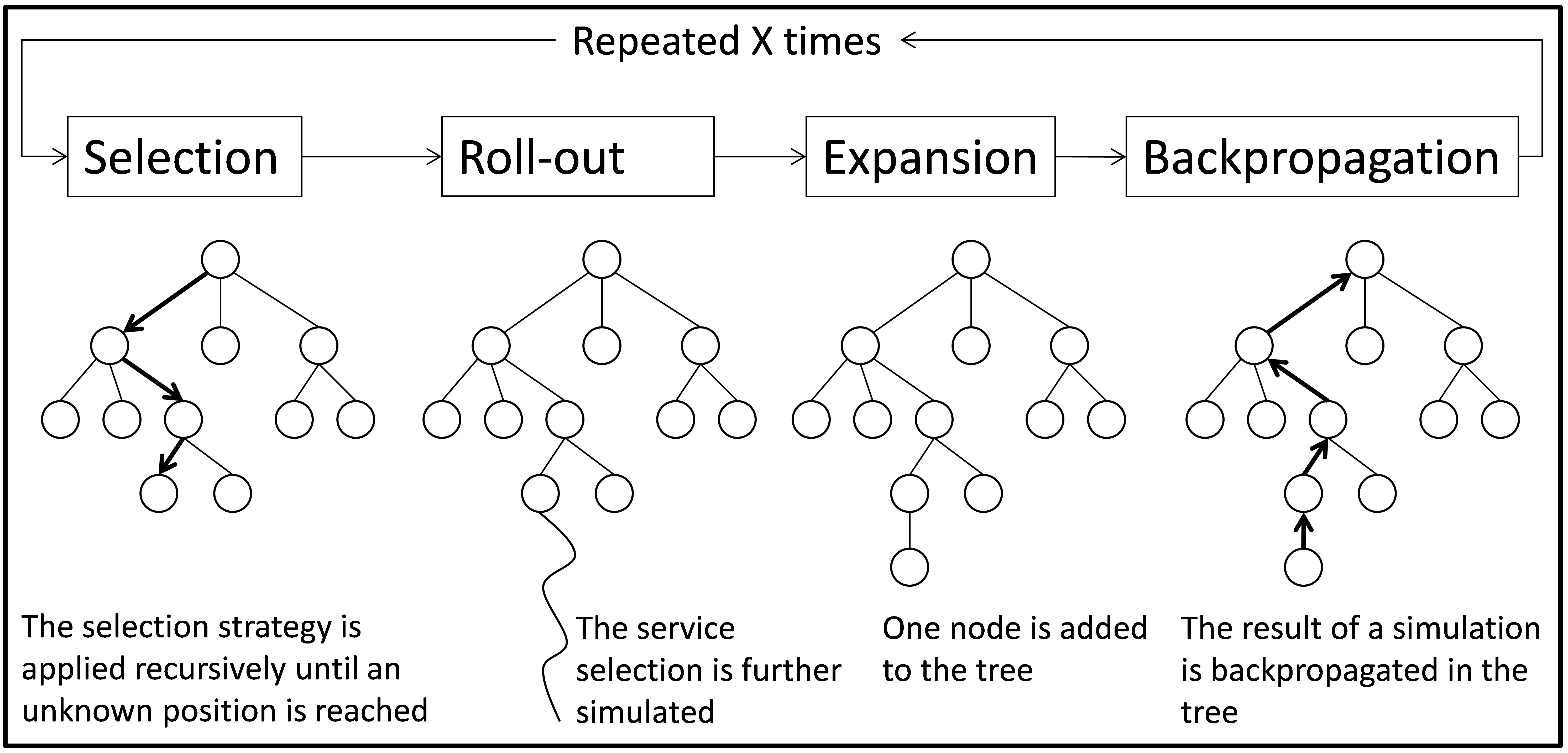}
  \caption{The four phases of Monte-Carlo Tree Search.}
  \label{fig:fig7}
\end{figure}

\subsubsection{Implementing MCTS for service selection}
When using MCTS for service selection, a few small adjustments are in order. Because service selection is essentially a single-agent search, MCTS is allowed to return an action sequence of care services instead of a single action after the search has ended. Furthermore, the roll-out phase of MCTS consists of randomly selecting care services that have not been selected yet in the current care plan. In the context of service selection, there are no actions that terminate the simulation. Therefore, a simulation can only end in a terminal state if no further actions are allowed, which is determined by the previously discussed limit of forming care plans of at most 8 services. The performance measure used at the end of a simulation is the risk of emergent care, which is calculated by the predictive model discussed in Section 2.2. The risk scores returned by the model are mirrored such that MCTS can maximize the rewards obtained during simulations. The reward values range from 0 to 1, where 1 indicates an ideal scenario with no risk of emergent care and 0 indicates a guaranteed emergent care. The MCTS configuration described above is henceforth referred to as vanilla MCTS.

\subsubsection{Enhancements}
As discussed, the state space is too large to traverse completely in reasonable time limits. Although MCTS provides a way of coping with this, the majority of search time will still be spent on generating the top part of the tree, while only a fraction of the bottom part of the tree is explored and taken into account by Monte-Carlo simulations. Therefore, only the first few care services of a care plan returned by MCTS have been thoroughly considered and evaluated during the search. The remaining chosen services of the plan were coincidentally found to be a good addition during one of the simulations. The selection of the last few services of a plan can however be as important as the selection of the first few services.
To circumvent this problem, two domain-independent MCTS enhancements for service selection were used: time-controlled MCTS and Progressive History/MAST. Both are discussed briefly in the subsections below.

\paragraph{Time-controlled MCTS.}
In order to prioritize the selection of each service in a care plan equally, the search should be spread out across the full depth of the search tree. Time-controlled MCTS [22] provides a domain-independent way of doing this by dividing the allowed search time by the number of actions that may be performed. Time-controlled MCTS starts by performing a search for selecting the first service of the care plan, with an allowed search time of \(t=\frac{T}{d}\) where \(T\) is the total allowed search time and \(d\) is the number of services to select. Once the first service is selected, the generated tree is kept in memory and used again during the selection of the second care service. However, the tree node corresponding to the first selected care service becomes the new root node. Therefore, the subsequent search for the selection of the second care service will start from the node of the previously chosen service. This process continuous until a care plan is formed or until no better plan is found. Figure 8 displays a tree generated by time-controlled MCTS in which two care services have already been chosen during the last 20 seconds. By setting the root node one level lower in the tree, the other available parts of the tree at that level are never evaluated by MCTS again. While obviously a greedy approach, this effectively prevents the search tree from becoming too large and allows MCTS to reach the lowest depth of the search tree. As a result, the last remaining options of a care plan are no longer chosen based on Monte-Carlo simulations, but are represented by a node in the tree.

% FIGURE 8
\begin{figure}
  \centering
  \includegraphics[width=0.9\textwidth]{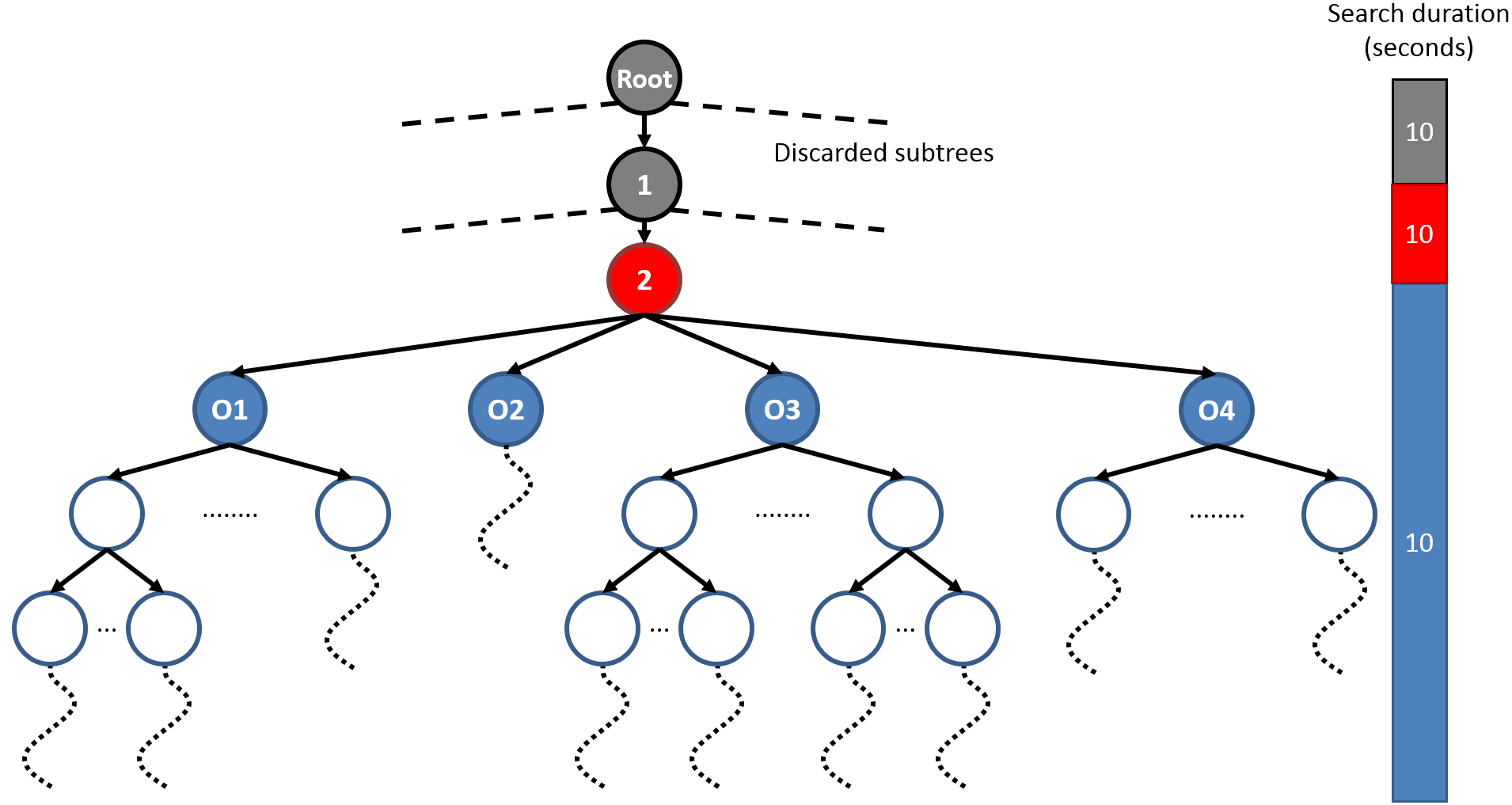}
  \caption{Tree structure after 20 seconds of search using time-controlled MCTS. Two out of the 8 services have already been chosen, while the next 10 seconds will be spent on choosing the third service, choosing from O1, O2, O3 or O4, marked in blue. Note that the new root node of the tree, indicated in red and marked by `2', is set to the second chosen service.}
  \label{fig:fig8}
\end{figure}

\paragraph{Progressive History and MAST.}
Progressive History (PH) [23] is applied to enhance the default UCT selection strategy. UCT may be unreliable when nodes have not been visited often. The general idea of PH is that actions that have been effective in general might be good candidates for selection when there is no clear winner among all the children of a node. \\ 

PH stores for each action a corresponding relative history score in a table. These scores are updated during the backpropagation phase of MCTS for each chosen action. By keeping track of the number of times \(hn_m\) that an action m was played and the resulting total reward \(hr_m\), the relative history score can be computed by \(\frac{hr_m}{hn_m}\). The relative history score is subsequently used by the PH formula during the selection phase, shown below in Formula 2.

\begin{equation}
    v_i=\frac{r_i}{n_i} +C\sqrt{\frac{\ln{n_p}}{n_i}}+\frac{hr_m}{hn_m} \times \frac{W}{(1-\frac{r_i}{n_i}) n_i+1)}
\end{equation}

In this formula, the variables \(v_i\), \(r_i\), \(n_i\), \(n_p\) and parameter \(C\) are identical to those in the original UCT formula [15]. Finally, \(W\) is a constant regulating the influence of the relative history score on the selection procedure. The relative history score will have a large influence when the corresponding node has not been visited often. Once the node has been visited a few times, the influence of the relative history score will gradually diminish until the selection strategy functions identical to the UCT strategy.\\

The relative history scores of the PH table are also used for an $\epsilon$-greedy MAST simulation strategy [24], which is based on the Move-Average Sampling Technique (MAST) proposed by [25]. This simulation strategy consists of performing the best available action, according to the heuristic scores of the PH table, with probability \(1-\epsilon\). A random action is executed with a small probability \(\epsilon\) (0.1 in this case) to prevent the simulations of becoming too deterministic. This enables MCTS to use PH scores in the roll-out phase, which might result in more accurate evaluations of roll-outs.

\section{Experimental setup}
MCTS was implemented in Java, while the predictive ensemble model was created in the R data analytics software package [26]. The model was converted to PMML format [27] in order to ensure compatibility with the Java environment. All experiments were performed on an AMD 3.4 GHz processor. If not stated otherwise, a search time of 60 seconds was allowed and a combination of 8 services should be selected by the automated service selection, for a given patient. The vanilla MCTS configuration achieved 960 simulations per second on average. \\

Unless mentioned specifically, each experiment consists of performing service selection for an identical set of 100 distinct patients, stratified over risk deciles. For each risk decile, 10 patients are randomly selected from the group of eligible patients. The risk score of a patient is determined by the developed predictive model when using the services recommended by the clinician in the dataset. This approach ensures that patients of various risk levels are equally represented in the test data. This also allows the inspection of service selection performance among low, medium and high risk patients. Furthermore, these 100 patients have not been included as training data for fitting the predictive model. The resulting test set with corresponding risk values is shown in Figure 9. The red lines indicate the boundaries of each decile. Note how the higher deciles cover larger ranges of risk. \\

Unless states otherwise, the performance measure is the average risk reduction. Risk reduction is defined by the difference between initial risk and resulting risk. The initial risk is assessed by the predictive model using the care plan originally recommended by the clinician in the NHHCS dataset. The resulting risk is assessed by the predictive model using the care plan recommended by MCTS. \\

The difference between initial and resulting risk, measured in percentage points, indicates the reduced risk per patient. The larger the risk reduction, the better the automated service selection is performing. Average risk reductions can be computed over the complete test set and per decile.\\

Most of the experiments discussed in this section were also performed using a non-weighted model, created by fitting logistic regression models on the NHHCS dataset without using the propensity score weights, but using an identical method of feature selection. Performance of service selection using this model was generally worse compared to using the weighted model. Moreover, manual inspection of a few cases showed that the selected services seemed far more random and incoherent with the patient characteristics, which is to be expected when not removing selection bias.

% FIGURE 9
\begin{figure}
  \centering
  \includegraphics[width=0.9\textwidth]{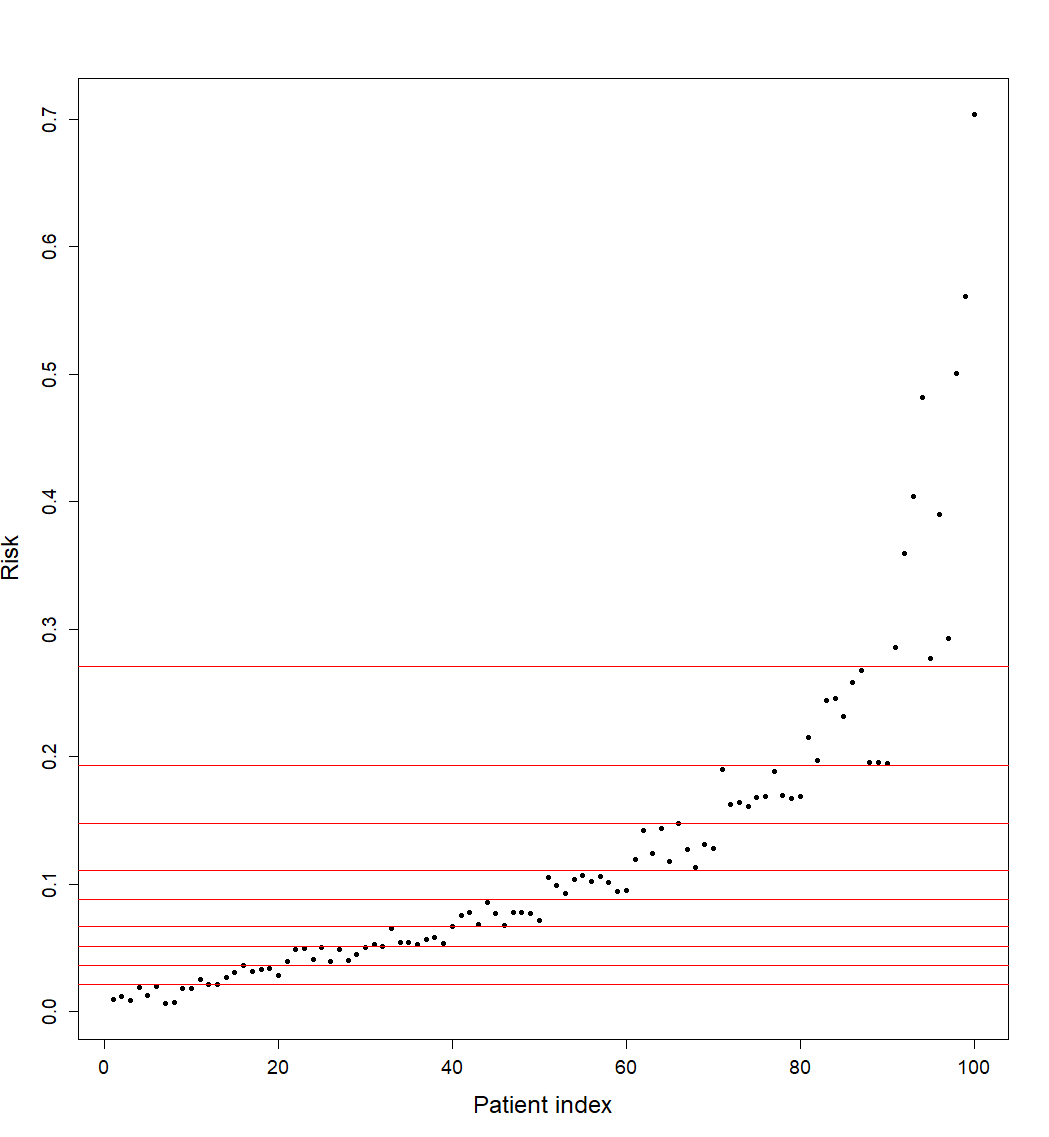}
  \caption{The test set used for the experiments, consisting of 100 patients chosen from a set of 500 randomly selected patients, stratified by risk deciles.}
  \label{fig:fig9}
\end{figure}

\section{Results}

\subsection{Predictive Model Evaluation}
The resulting ensemble model consists of 15 logistic regression models. The receiver operating characteristic (ROC) curve [28] is shown in Figure 10, indicating sensitivity and specificity using 10-fold cross-validation at differing thresholds of classification. The model achieves an area under curve (AUC) of $76 \pm 2\%$. Although validation of the model is warranted, this is a promising result. The ROC curves of the individual logistic regression models of the ensemble model are indicated in Figure 11, with AUC values ranging from $69 \pm 3\%$ to $71 \pm 2\%$, which indicates that the ensemble approach adds a significant amount of predictive performance.\\

% FIGURE 10
\begin{figure}
  \centering
  \includegraphics[width=0.9\textwidth]{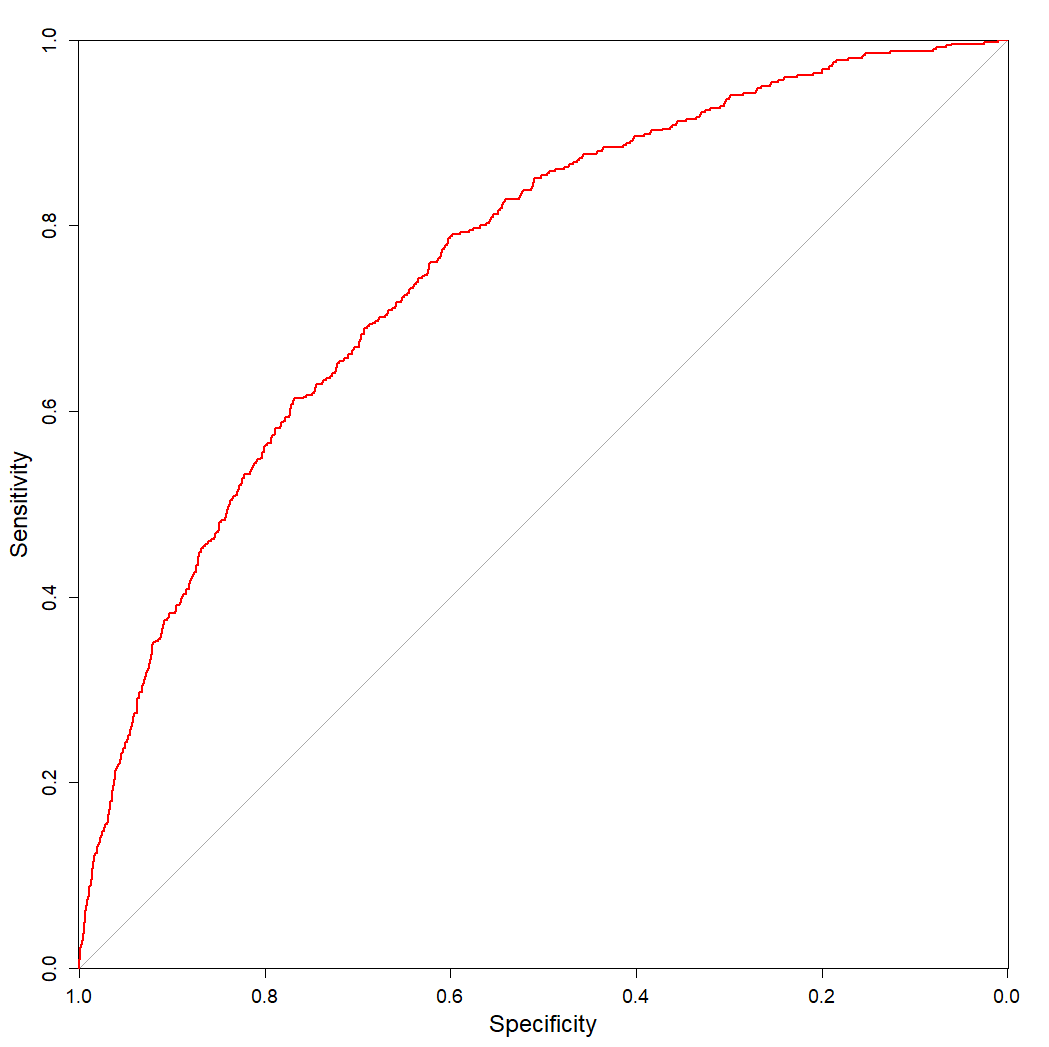}
  \caption{ROC curve of the ensemble model.}
  \label{fig:fig10}
\end{figure}

% FIGURE 11
\begin{figure}
  \centering
  \includegraphics[width=0.9\textwidth]{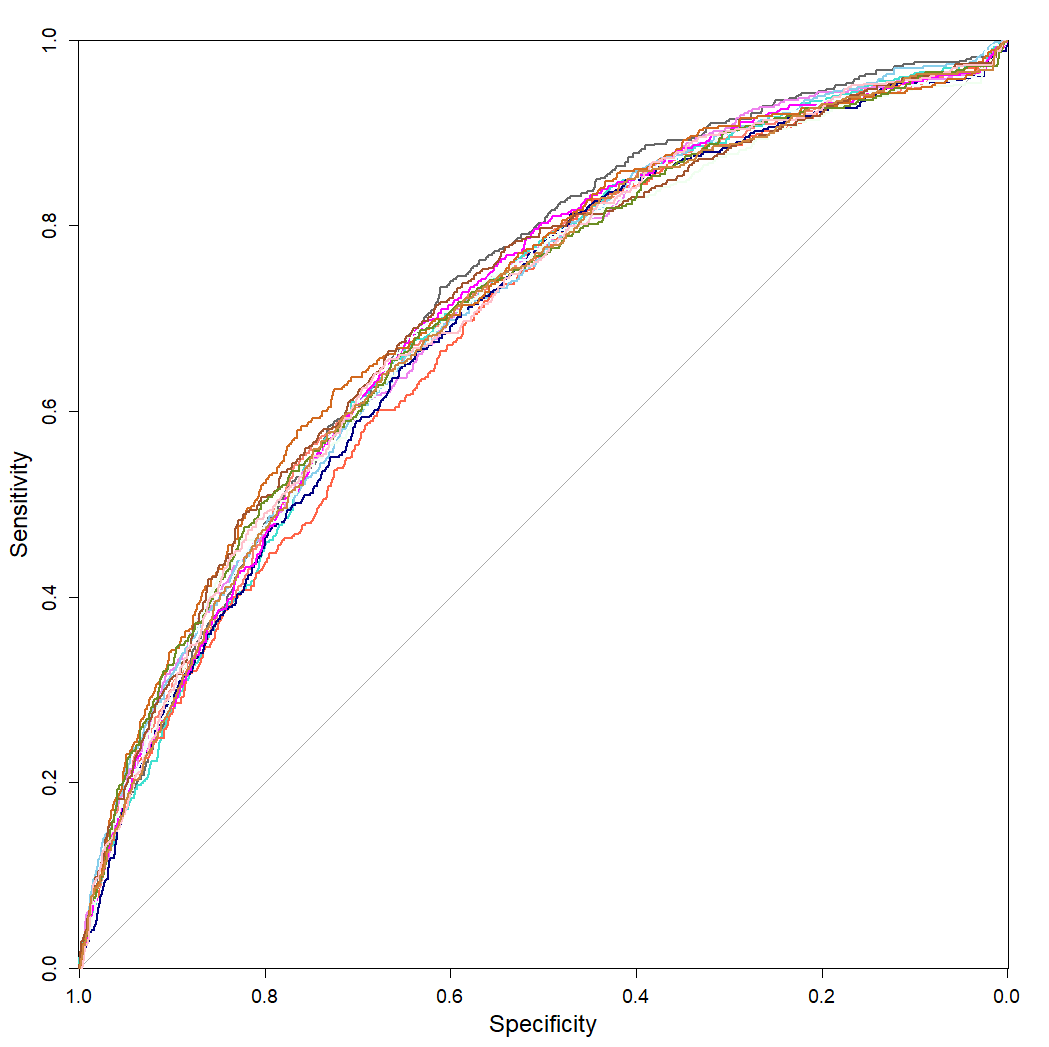}
  \caption{ROC curves of the 15 models forming the ensemble model.}
  \label{fig:fig11}
\end{figure}

The individual models of the ensemble consist of 35 predictors on average, with a standard deviation of 3.6. The ensemble model contains 529 predictors in total, each one corresponding to an interaction between a care service and either a patient characteristic or another care service. A care service is represented in 7 interactions on average with a standard deviation of 3.7. The most and least frequent care services are mentioned in Table 3.

% TABLE 3
\begin{table}[]
\begin{center}
\caption{Most and least frequent care services represented in predictors of the ensemble model.}
\label{tab:my-table3}
\begin{tabular}{|l|l|l|l|}
\hline
\textbf{Most frequent} &  & \textbf{Least frequent} &  \\ \hline
\textbf{Service} & \textbf{Frequency} & \textbf{Service} & \textbf{Frequency} \\ \hline
Skilled nurse & 15 & Dietary counselling & 2 \\ \hline
Dealing with difficult behaviors services & 14 & Respite services & 2 \\ \hline
Medication management & 14 & Continuous positive pressure airway & 1 \\ \hline
Oxygen concentrator & 14 & Interpreter services & 1 \\ \hline
Wound care & 14 & Podiatry services & 1 \\ \hline
\end{tabular}
\end{center}
\end{table}

Manual inspection of the model shows several noteworthy interactions. For example, one of the models contains the interaction between abnormality of gait and transport services. Abnormality of gait indicates that a patient is having difficulties with walking or keeping balance, which does not allow them to move freely or drive a car. This interaction indicates that the presence of the corresponding ICD9 code combined with the usage of transport services negatively influences the probability of emergency readmission. Another example is the usage of a glucose reading device and the ICD9 code indicating a urinary tract infection. Medical literature reports that diabetes patients are more likely to have an urinary tract infection, due to the consistent presence of high amounts of glucose in urine [29]. The developed predictive model indicates that using a glucose reader device when having a urinary tract infection lowers the probability of emergent care.

\subsection{Parameter tuning}
Before any other experiments can be conducted, the $C$ parameter of UCT and subsequently the $W$ parameter of PH/MAST need to be tuned. The $C$ parameter is tuned by trial and error, measuring performance with $C$ values ranging from 0 to 100. The best performance is achieved when using a $C$ value of 0.05, shown in Table 4, in which the average reduced risk for each decile is indicated. Column 1 represents the lowest decile, 10 indicates the highest decile. The performance for the complete test set is indicated in the rightmost column.

% TABLE 4
\begin{table}[]
\begin{center}
\caption{Average reduced risk for various tuning configurations per decile and over the complete test set.}
\label{tab:my-table4}
\begin{tabular}{|l|l|l|l|l|l|l|l|l|l|l|l|}
\hline
\textbf{Configuration} & \textbf{1} & \textbf{2} & \textbf{3} & \textbf{4} & \textbf{5} & \textbf{6} & \textbf{7} & \textbf{8} & \textbf{9} & \textbf{10} & \textbf{All} \\ \hline
$C=0.05$ & 1.15 & 2.59 & 4.37 & 5.38 & 7.17 & 9.57 & 12.45 & 16.30 & 20.35 & 38.92 & $11.82 \pm 0.001$ \\ \hline
$W=0.1$ & 1.25 & 2.70 & 4.42 & 5.45 & 7.23 & 9.63 & 12.45 & 16.39 & 20.39 & 38.93 & $11.88 \pm 0.012$ \\ \hline
$W=0$ & 1.23 & 2.66 & 4.43 & 5.46 & 7.22 & 9.66 & 12.43 & 16.32 & 20.30 & 39.02 & $11.87 \pm 0.009$ \\ \hline
\end{tabular}
\end{center}
\end{table}

While the average reduced risk over all patients is 11.82 percentage points, the largest risk reductions are achieved in the highest deciles. The automated service selection manages to reduce the risk of patients in the highest decile by 38.92 percentage points on average. Lower decile patients are not as potent in this regard, as these patients already have a low risk before applying MCTS service selection, either due to the services they are already receiving or the patient characteristics. The maximal reduction of risk is achieved in the $10^{th}$ decile, in which a patient went from a $70.43\%$ to a $13.89\%$ risk using MCTS service selection, which corresponds to a decrease of 56.54 percentage points. The smallest reduction took place in the first decile, going from a $0.76\%$ risk to a $0.23\%$ risk. For each of the following experiments, the $C$ parameter in the UCT formula was set to 0.05.\\

The fact that a $C$ value of 0.05 achieves the best result is remarkable, as this means that MCTS largely favors exploitation over exploration. This indicates that the values returned by the roll-out phase are good indicators of the current care plan. Moreover, a greedy approach by not exploring other nodes in the same turn is also partially allowed due to the structure of the search. Because service selection is a single-agent search scenario, any of the options not chosen during the current turn may still be chosen in the next turn.\\

The $W$ parameter of the Progressive History/MAST (PH) enhancement was tuned in a similar fashion, with values ranging from 0 to 1000. The highest average reduced risk is achieved with a $W$ value of 0.1, resulting in an average reduced risk of 11.88 percentage points, indicated in Table 4. The highest decile of patients has an average risk reduction of 38.93 percentage points. When service selection is performed with a $W$ value of 0, the PH/MAST enhancement is only active in the roll-out phase of MCTS. This configuration is also shown in Table 4. The results indicate an improvement in performance compared to the vanilla configuration, as the average risk reduction increases from 11.82 to 11.87 percentage points. For any of the following experiments that included PH/MAST, $W$ was set to 0.1.

\subsection{Performance of enhancements}
Next, the performance of the enhancements can be investigated. Service selection is performed with varying search times and combinations of enhancements, more specifically, a vanilla MCTS configuration, MCTS combined with PH/MAST, time-controlled MCTS and time-controlled MCTS with PH/MAST. The average risk reductions over the complete test set are shown in Table 5.

The results show that performance improves for each MCTS configuration as longer search times are allowed. The improvements in performance are however small, suggesting that quickly finding a decent care plan is not that difficult, but finding the optimal or close to optimal solution is quite difficult. Both enhancements outperform the vanilla configuration at every search time, but PH/MAST performs slightly better than time-controlled MCTS. It might however be the case that time-controlled MCTS outperforms PH when larger care plans are allowed, as this yields deeper trees. The combination of the two enhancements results in the highest average risk reduction overall for any search time. The best performing configuration is the 60 second time-controlled MCTS with PH/MAST and was therefore used for further experiments. \\

It should be noted that the enhancements are especially useful when search time is limited, as the largest relative improvement in performance by using the enhancements is achieved when only short search times are allowed. This is likely the case in more practical, real-world settings. 

% TABLE 5
\begin{table}[]
\begin{center}
\caption{Average reduced risk for several MCTS configurations and varying search times.}
\label{tab:my-table5}
\begin{tabular}{|l|l|l|l|l|}
\hline
\textbf{Search Time (seconds)} & \textbf{Vanilla} & \textbf{PH/MAST} & \textbf{Time-controlled} & \textbf{PH \& Time-controlled} \\ \hline
1 & $10.19 \pm 0.065$ & $10.98 \pm 0.022$ & $10.59 \pm 0.118$ & $11.01 \pm 0.100$ \\ \hline
5 & $11.09 \pm 0.060$ & $11.52 \pm 0.017$ & $11.36 \pm 0.110$ & $11.69 \pm 0.008$ \\ \hline
10 & $11.40 \pm 0.007$ & $11.70 \pm 0.038$ & $11.53 \pm 0.001$ & $11.81 \pm 0.020$ \\ \hline
20 & $11.60 \pm 0.044$ & $11.80 \pm 0.008$ & $11.72 \pm 0.007$ & $11.83 \pm 0.016$ \\ \hline
40 & $11.74 \pm 0.025$ & $11.81 \pm 0.051$ & $11.83 \pm 0.018$ & $11.89 \pm 0.022$ \\ \hline
60 & $11.79 \pm 0.029$ & $11.88 \pm 0.012$ & $11.84 \pm 0.001$ & $11.89 \pm 0.014$ \\ \hline
\end{tabular}
\end{center}
\end{table}

\subsection{Varying the number of services}
Although a patient receives 8 care services on average in the NHHCS dataset, there is quite a large spread in care plan size. Severe cases will need more than 8 services, while more stable cases will require less than 8 services. The next experiment investigates how the average risk reduction changes when size of the care plans is altered. The results are displayed in Table 6.\\

Performance steadily increases when larger care plans are allowed. However, this effect does have diminishing returns, as the performance at combinations of size 30 seems to be stagnating. When only allowed to recommend one care service, MCTS still achieves an average risk reduction of 8.21 percentage points. This is quite remarkable, as this shows even a single care service can reduce the risk of re-hospitalization significantly. However, larger risk reductions are achieved for larger care plans, especially among the higher deciles.

% TABLE 6
\begin{table}[]
\begin{center}
\caption{Average reduced risk per decile and over the complete test set for varying care plan sizes.}
\label{tab:my-table6}
\begin{tabular}{|l|l|l|l|l|l|l|l|l|l|l|l|}
\hline
\textbf{Nr. Of Services} & \textbf{1} & \textbf{2} & \textbf{3} & \textbf{4} & \textbf{5} & \textbf{6} & \textbf{7} & \textbf{8} & \textbf{9} & \textbf{10} & \textbf{All} \\ \hline
1 & 0.66 & 1.61 & 2.71 & 3.47 & 4.68 & 7.05 & 8.67 & 12.85 & 13.43 & 27.02 & $8.21 \pm 0.01$ \\ \hline
5 & 1.12 & 2.44 & 4.08 & 5.02 & 6.65 & 9.08 & 11.67 & 15.48 & 18.68 & 36.38 & $11.06 \pm 0.015$ \\ \hline
8 & 1.27 & 2.72 & 4.49 & 5.46 & 7.21 & 9.68 & 12.49 & 16.39 & 20.32 & 38.86 & $11.89 \pm 0.014$ \\ \hline
10 & 1.28 & 2.81 & 4.52 & 5.48 & 7.47 & 9.94 & 12.68 & 16.62 & 20.83 & 39.71 & $12.14 \pm 0.019$ \\ \hline
15 & 1.32 & 2.94 & 4.75 & 5.79 & 7.69 & 10.14 & 12.91 & 16.97 & 21.53 & 40.55 & $12.46 \pm 0.020$ \\ \hline
20 & 1.35 & 2.90 & 4.84 & 5.71 & 7.78 & 10.28 & 13.15 & 17.03 & 21.66 & 41.35 & $12.61 \pm 0.003$ \\ \hline
30 & 1.36 & 3.01 & 4.80 & 5.90 & 7.84 & 9.67 & 13.06 & 17.15 & 21.93 & 40.96 & $12.57 \pm 0.089$ \\ \hline
\end{tabular}
\end{center}
\end{table}

\subsection{Comparison with Dijkstra}
MCTS has shown to be capable of significantly reducing the risk of re-hospitalization for the provided test set. However, the predictive model can also be used by other search techniques. Because there is no admissible heuristic function for service selection, A* is not applicable.  A suitable alternative is the Dijkstra algorithm [30]. Dijkstra can be used in this context because the service selection problem can be defined by a graph. Vertices in the graph represent care plans, allowing a selection of exactly 8 services.\\

The distance between two vertices is calculated as shown in Formula 3, in which $d_{i,j}$ is the distance between vertices $i$ and $j$ and $risk_k$ indicates the risk score of the patient when using the care plan corresponding to vertex $k$.

\begin{equation}
    d_{i,j} = 1-(risk_i-risk_j)
\end{equation}

The performances of Dijkstra and MCTS are compared by letting each algorithm perform service selection on the same test set and allowing search times of 60 seconds. The results are shown in Table 7. MCTS performs significantly better than Dijkstra, as MCTS achieves an average risk reduction of 11.88 percentage points compared to the 10.02 of Dijkstra. MCTS consistently achieves better risk reductions in each decile. Although the difference in average risk reduction for all patients is small, the largest differences are obtained in the higher deciles, in which MCTS achieves an average risk reduction of 38.93 percentage points, while Dijkstra only reaches a reduction of 33.03 percentage points. Therefore, MCTS outperforms Dijkstra.

% TABLE 7
\begin{table}[]
\begin{center}
\caption{Average reduced risk using either MCTS or Dijkstra, per decile and over the complete test set.}
\label{tab:my-table7}
\begin{tabular}{|l|l|l|l|l|l|l|l|l|l|l|l|}
\hline
\textbf{Algorithm} & \textbf{1} & \textbf{2} & \textbf{3} & \textbf{4} & \textbf{5} & \textbf{6} & \textbf{7} & \textbf{8} & \textbf{9} & \textbf{10} & \textbf{All} \\ \hline
MCTS & 1.25 & 2.70 & 4.42 & 5.45 & 7.23 & 9.63 & 12.45 & 16.39 & 20.39 & 38.93 & 11.88 \\ \hline
Dijkstra & 0.98 & 2.15 & 3.57 & 4.46 & 5.89 & 8.32 & 10.58 & 14.49 & 16.74 & 33.03 & 10.02 \\ \hline
\end{tabular}
\end{center}
\end{table}

\subsection{Single case evaluation}
This section takes a closer look at how the selected care plans for low, medium and high risk patients relate to the characteristics of the patients. More specifically, each case discussed below aims to relate the recommended care services to one or more patient characteristics. These relations are supported by medical literature if possible, otherwise common-sense knowledge is used. Hence, some suggested relations are speculative. Another method for investigating these relations is to inspect the relevant interactions in the ensemble model. This will however result in a far too detailed and unnecessarily complicated analysis for the scope of this section.\\

The selected care plans for a low, medium and high risk patient are discussed in Subsections 4.6.1, 4.6.2 and 4.6.3, respectively.

\subsubsection{Case 1}
A low risk case is displayed in Table 8. The patient’s conditions are shown in the leftmost column. The resulting MCTS selection is shown in the middle column, corresponding to the order in which MCTS chose these services during the search. The selection of the clinician is shown in the rightmost column. The numbers stated after a care service refer to the patient’s conditions and therefore indicate the suggested relations between characteristics and care services. The clinician’s plan achieves a risk of $3.4\%$. The MCTS care plan achieves a risk of $0.7\%$, by keeping only two of the clinician’s recommended services and adding 6 different care services. \\

The first choice of MCTS is `ORTHOTIC', which indicates the use of orthoses. This is most likely recommended due to the diagnosis of diabetes, as diabetics are known for using orthotic treatment to prevent recurrent diabetic foot ulcers [31]. Service `PARENTIV' indicates the usage of parenteral nutrition. Previous research shows a positive effect on anemia patients when provided with parenteral nutrition [32]. Dietary services, indicated by `DIETSRV', is most likely provided because the patient has diabetes, a condition that is usually caused by an unhealthy diet [33]. Moreover, the fatigue and malaise of the patient might also be treated by this service, as fatigue and malaise can be caused by a deficient diet. MCTS also selects `APNEA', a device that monitors the user while sleeping to detect pauses in breathing. It is interesting that this service is recommended, as the patient has atrial fibrillation, a heart rhythm disorder. Sleep apnea is prevalent among patients with atrial fibrillation [34]. Moreover, previous work shows that $46\%$ of malaise and fatigue patients have disruptive sleep apnea [35,36]. Furthermore, it is also known that there is a direct relation between diabetes patients and sleep apnea [37,38]. \\

The remaining care services are not explicitly supported by literature. The `PHARMACY' option is most likely selected because the patient has several conditions that are usually treated with drugs. The `ADLSRVCE' indicates help from the agency with bathing, dressing, toileting or feeding. Furthermore, `ABUSIVE' represents a service that investigates cases of abuse or neglect of patients. Both options are most likely related to the malaise and fatigue of the patient. Finally, the recommendation of `EQUIPMNT' indicates that the patient will be receiving instructions for using the provided devices, which in this case is the apnea monitor.

% TABLE 8
\begin{table}[]
\begin{center}
\caption{Service selection results for a low risk patient.}
\label{tab:my-table8}
\begin{tabular}{|lll|}
\hline
\textbf{Conditions} & \textbf{Selection MCTS} & \textbf{Selection Nurse} \\
\hline
1 Diabetes & ORTHOTIC (1) & WLKCANE \\
2 Hypertension & PHARMACY (2, 3, 6) & SHOWBAT \\
3 Atrial fibrillation & PARENTIV (5, 4) & PHYTHERA \\
4 Malaise and fatigue & DIERTSRV (1, 4, 2) & MEDICATE \\
5 Anemia & ADLSRVCE (4) & SAFETY \\
6 Urinary Tract Infection & ABUSIVE (4) & EQUIPMNT \\
 & APNEA (3, 4, 1) & ADLSRVCE \\
 & EQUIPMNT (3, 4, 1) &  \\
 \hline
\textbf{Number of Services:} & 8 & 7 \\ \hline
\textbf{Risk:} & 0.7\% & 3.4\% \\
\hline
\end{tabular}
\end{center}
\end{table}

\subsubsection{Case 2}
The second case, shown in Table 9, is a medium risk patient with an initial risk of $10.6\%$ when using the care services recommended by the clinician. When using the care plan selected by MCTS, the risk of emergency re-hospitalization drops to $0.6\%$, while also using only 8 services instead of 11. \\

MCTS starts by choosing the apnea monitor service. Both CHF and COPD are common co-morbidities of apnea [39,40]. Moreover, when a patient has both COPD and sleep apnea, also known as the overlap syndrome, there is a greater risk of heart failure, which results in higher morbidity rates overall [41].  Therefore, it is recommended to actively search for this condition among COPD patients. MCTS most likely chooses to treat the COPD condition with respiratory therapy and pharmacological treatment, indicated by `RESTHERA' and `PHARMACY'. In order to cope with the abnormality of gait, several services are chosen: ‘TRANEQUP' and `BEDTABLE'. The first indicates the provision of transport equipment (e.g. stair lift, gait belt). The second choice indicates that a bed table is provided, which is remarkable, as this service is a quality of life service which might not be expected to directly reduce the risk of emergency re-hospitalization. MCTS continues by appropriately selecting `EQUIPMNT' and `DEVSPPRT', which indicates that the agency will be instructing and helping the patient with the installation of these devices and monitoring the usage. The `CAM' option represents the usage of complementary and alternative medicine. This cannot be directly related to any of the patient’s conditions.

% TABLE 9
\begin{table}[]
\begin{center}
\caption{Service selection result for a medium risk patient.}
\label{tab:my-table9}
\begin{tabular}{|lll|}
\hline
\textbf{Conditions} & \textbf{Selection MCTS} & \textbf{Selection Nurse} \\
\hline
1 COPD & APNEA (1, 2) & WLKCANE \\
2 CHF & RESTHERA (1) & CHAIRS \\
3 Abnormality of gait & BEDTABLE (3) & GRABBARS \\
 & CAM & SHOWBATH \\
 & EQUIPMNT (1, 2, 3) & SKILLED \\
 & DEVSPPRT (1, 2, 3) & OCCTHERA \\
 & TRANEQUP (3) & PHYTHERA \\
 & PHARMACY (1, 2) & MEDICATE \\
 &  & ADLSRVCE \\
 &  & SAFETY \\
 &  & EQUIPMNT \\ \hline
\textbf{Number of services:} & 8 & 11 \\ \hline
\textbf{Risk:} & 0.6\% & 10.6\% \\ \hline
\end{tabular}
\end{center}
\end{table}

\subsubsection{Case 3}
Table 10 shows the results for a high risk patient. The patient has a remarkably high risk of $57.8\%$ by using the 7 care services selected by the clinician. The MCTS plan of 8 care services, however, reduces the risk to $3.8\%$. This is an example of how high risk patients have the largest potential for significant risk reductions.\\

MCTS starts by choosing `EATDEVIC', indicating that the patient will be receiving specialized eating devices (e.g. non-spill cup, deeper plates and bowls to prevent spilling, specialized cutlery). The specialized cutlery usually has larger and softer grips and are angled such that the patient does not have to twist his or her wrists when eating. This service is most likely selected for the arthropathy of the patient. Next, MCTS recommends `WOUNDS', which is a service that provides wound care. Some variations of arthropathy, such as neuropathic arthropathy, are known for causing chronic infections and ulcerations of joints and surrounding skin [42]. The following recommendation, `PHARMACY', is most likely selected for COPD and anemia medication. MCTS subsequently selects `EQUIPMNT' to, most likely, instruct the patient with the eating devices and apnea monitor. Orthotic services are selected next, probably to ease the muscle pains and aches caused by the arthropathy. Next, respiratory services are included almost certainly to cope with the breathing difficulties caused by COPD. Following that is the selection of an apnea monitor, which is probably chosen due to the COPD diagnosis as discussed in case 2. Finally, the care plan is completed by the recommendation of regularly paying a visit to a physician, which may be related to any of the chronic conditions.

% TABLE 10
\begin{table}[]
\begin{center}
\caption{Service selection result for a high risk patient.}
\label{tab:my-table10}
\begin{tabular}{|lll|}
\hline
\textbf{Conditions} & \textbf{Selection MCTS} & \textbf{Selection Nurse} \\ \hline
1 COPD & EATDEVIC (2) & ENTEROST \\
2 Arthropathy & WOUNDS (2) & DEVSPPRT \\
3 Anemia & PHARMACY (1, 3) & OSTOMY \\
 & EQUIPMNT (1, 2) & SKILLED \\
 & ORTHOTIC (2) & DIETARY \\
 & RESPSRVC (1) & DIETCOUN \\
 & APNEA (1) & WOUNDS \\
 & PHYSICIAN &  \\ \hline
\textbf{Number of services:} & 8 & 7 \\ \hline
\textbf{Risk:} & 3.8\% & 57.8\% \\ \hline
\end{tabular}
\end{center}
\end{table}

\subsubsection{Findings}
These cases show that MCTS is capable of tailoring care plans to a patient’s conditions. Examples are the selection of specialized eating devices for arthropathy, orthotic services when diagnosed with diabetes, or choosing an apnea monitor when having both COPD and CHF. Moreover, these specific recommendations are immediately selected in the root of the search tree as the first service for the care plan. This shows that the predictive model is not only capable of generalizing the frequent relations expressed in the NHHCS data, but also encapsulates the less frequent but equally important relations. \\

It should however be noted that these recommendations might also be too specific or just not applicable for a patient. To illustrate this point, in the third case, a high risk patient receives eating devices to presumably deal with arthropathy in the hands of the patient. However, it is not specifically known that this disease is located in the hands of the patient. Because the model generalized that patients with arthritis usually benefit from eating devices, this is recommended to the patient. A higher granularity in data features could separate these cases (i.e. separate features indicating where the patient has arthropathy), but it would most likely be unfeasible to create such a thorough description to circumvent this issue altogether. \\

An even more important observation is that the nurse seems to be selecting options that improve the quality of life of patients, or those that are covered by insurance, instead of reducing the risk of emergency re-hospitalization. This could explain the differing plan selections and the large reductions in risk. This does however still show that there is plenty of room for improvement when aiming for a reduction in emergency readmission rates. 

\section{Discussion}
\subsection{Summary, achievements and limitations}
The goal of this research was to create an automated post-acute service selection process for addressing some of today’s most pressing challenges in the healthcare domain, specifically insufficient treatment efficacy and high treatment costs, caused by a lack of tailoring to the patient and the absence of applying empirical data on treatment efficacy in practice.\\

The discussed process tackles these challenges by applying predictive modeling based on a set of historical patient treatment and effects, in order to model the treatment effect of a combination of post-acute services given the patient’s characteristics, such as comorbidities and medical procedures. The predictive model is subsequently used as a utility function in Monte-Carlo Tree Search in order to find combinations of services that maximize the risk reduction indicated by this utility function. \\

The results show a significant reduction in risk of re-hospitalization compared to the original selection made by the clinician, given that approximately 12 percentage points in risk reduction is achieved on average over the complete test population. Higher reductions of roughly 40 percentage points on average are observed in the highest risk categories. Moreover, individual case analyses showed that the recommendations are often relatable to one or several comorbidities of the patient.\\

Some limitations are however encountered as well. The data must be sufficiently expressive to capture the often crucial details for tailored service selection. This is not always the case, leading to services being selected that are not applicable for the patient in question. Another limitation is that the current proposed technique does not take the aspect of time into account: it assumes all services will be provided simultaneously for an undetermined period of time. Moreover, the automated process may target a different goal than the clinician (e.g. risk reduction re-hospitalization versus coverage by insurance), which leads to conflicting interests and therefore contradicting service selection. \\

It is for these reasons that the developed process is most suited to work in collaboration with a clinician. In this scenario, the automated process would recommend an initial combination of post-acute services given a patient, making sure that none of the relevant options are accidentally overlooked by the clinician or that unknown effects of treatments on emergency re-hospitalization are not taken into account. The clinician can then alter the care plan as he or she finds fitting. This ensures that services deemed insignificant for the outcome of the predictive model but significant for the clinician’s goals are included as well. Additionally, it prevents that services are recommended that are not applicable for the patient.

\subsection{Future work}
While the proposed process already performs quite well, there is still room for improvement. A selection of interesting areas is shown below:

\begin{itemize}
    \item As the largest risk reductions are achieved in the highest decile, risk based stratification could be used for selectively applying automated service selection to high risk patients for optimal return on investment.
    \item Several search algorithms could be investigated instead of MCTS, as longs as they also use a utility function for assigning a quality to their current selection of services. An example would be using a genetic algorithm, using the predictive model for determining fitness scores. Another possibility is describing the service selection problem as a linear programming problem and using a linear solver to find good or even optimal combinations of services. Other variations of MCTS are also possible, such as Nested Monte-Carlo Tree Search [43].
    \item Several alternatives for the ensemble predictive model could be investigated, such as neural nets.
    \item It would be interesting to see how the performance changes when there is no limitation regarding the number of services in a combination. In that case, MCTS is free to choose a combination of three services for one patient, but a combination of 12 for another. This reflects the notion that the number of services varies per patient depending on their health situation. A penalty could be needed for including additional services ensuring that smaller sets of services are preferred over larger sets if the added risk reduction is small.
    \item The financial aspect of service selection should not be ignored. Usually, the options to be chosen from are limited by a total budget, or by what is covered by insurance. These restrictions could for example be incorporated into the rewards of MCTS, by limiting the depth of the tree by a total sum of incurred costs, or by removing options (i.e. child nodes) from the tree that are not covered by insurance.
    \item An advantage of using the approach of predictive models is that the goal of MCTS service selection can be easily changed by using a model with a different outcome, such as financial costs. Mixing several goals is also a possibility. In this way, a balance can be struck to select services which simultaneously, for example, provide quality of life and reduce risk of emergency re-hospitalization. These preferences can then also be altered by the user, by weighting the scores of the individual models. 
\end{itemize}

\subsection{Conclusion}
Applying automated service selection as proposed in this article may significantly reduce risk of re-hospitalization, thereby improving treatment efficacy and reducing readmission related healthcare costs. By using a predictive model that takes the characteristics of the patient into account for predicting this risk, the recommended care plan will be tailored to the patient and moves away from the “one-size fits all” mentality of current day medical treatment. As these recommendations are purely based on the empirical evidence obtained from historical admission related patient data, the proposed technique also shows how empirical evidence can be applied directly in the medical field. \\

These significant advantages, combined with a growing body of electronic patient records and AI techniques, it seems that there is enormous potential for improving service selection in the near future. This will allow clinicians to focus on what they do best: providing care to their patients.

\section*{Conflict of interest}
Cliff Laschet, Jorn op den Buijs and Steffen Pauws are employees of Royal Philips.

\section*{References}

\begin{enumerate}
\item K.H. Bowles, D.E. Holland, S.L. Potashnik, Implementation and Testing of Interdisciplinary Decision Support Tools to Standardize Discharge Planning, in: NI 2012 Proc. 11th Int. Congr. Nurs. Informatics, 2012: pp. 41–45.
\item D.E. Holland, M.R. Harris, C.L. Leibson, V.S. Pankratz, K.E. Krichbaum, Development and validation of a screen for specialized discharge planning services, Nurs. Res. 55 (2006) 62–71.
\item K.H. Bowles, A. Hanlon, D. Holland, S.L. Potashnik, M. Topaz, Impact of discharge planning decision support on time to readmission among older adult medical patients, Prof. Case Manag. 19 (2014) 29–38. doi:10.1097/01.PCAMA.0000438971.79801.7a.
\item C.C. Bennett, K. Hauser, Artificial intelligence framework for simulating clinical decision-making: A Markov decision process approach, Artif. Intell. Med. 57 (2013) 9–19. doi:10.1016/j.artmed.2012.12.003.
\item R. Mechanic, Post-Acute Care — The Next Frontier for Controlling Medicare Spending, N. Engl. J. Med. 370 (2014) 692–694. doi:10.1056/NEJMp1315607.
\item Centers for Disease Control and Prevention, NHHCS - National Home and Hospice Care Survey Homepage, (2007). https://www.cdc.gov/nchs/nhhcs/index.htm (accessed January 25, 2018).
\item A.L. Jones, L. Harris-Kojetin, R. Valverde, Characteristics and use of home health care by men and women aged 65 and over, Natl. Health Stat. Report. (2012) 1–7.
\item A. Bercovitz, A. Moss, M. Sengupta, E.Y. Park-Lee, A. Jones, L.D. Harris-Kojetin, An overview of home health aides: United States, 2007., Natl. Health Stat. Report. (2011) 1–31. doi:10.1002/hed.21840.
\item C. Caffrey, M. Sengupta, A. Moss, L. Harris-Kojetin, R. Valverde, Home health care and discharged hospice care patients: United States, 2000 and 2007, Natl. Health Stat. Report. (2011) 1–27.
\item A. Linden, J.L. Adams, Using propensity score-based weighting in the evaluation of health management programme effectiveness, J. Eval. Clin. Pract. 16 (2010) 175–179. doi:10.1111/j.1365-2753.2009.01219.x.
\item C.M. Bishop, Pattern Recognition And Machine Learning, 2006. doi:10.1117/1.2819119.
\item P. Pudil, J. Novovičová, J. Kittler, Floating search methods in feature selection, Pattern Recognit. Lett. 15 (1994) 1119–1125. doi:10.1016/0167-8655(94)90127-9.
\item P.E. Hart, N.J. Nilsson, B. Raphael, A Formal Basis for the Heuristic Determination of Minimum Cost Paths, Syst. Sci. Cybern. IEEE Trans. 4 (1968) 100–107.
\item R. Coulom, Efficient Selectivity and Backup Operators in Monte-Carlo Tree Search, in: Proc. 5th Int. Conf. Comput. Games 2006, Springer-Verlag, Berlin, Heidelberg, 2007: pp. 72–83. doi:10.1007/978-3-540-75538-8$\_$7.
\item L. Kocsis, C. Szepesvári, Bandit based Monte-Carlo Planning, in: Eur. Conf. Mach. Learn. ECML-06., Springer, 2006: pp. 282–293.
\item J. van Eyck, J. Ramon, F. Güiza, G. Meyfroidt, M. Bruynooghe, G. van den Berghe, Guided Monte Carlo Tree Search for Planning in Learned Environments, in: 5th Asian Conf. Mach. Learn. ACML 2013, 2013: pp. 33–47.
\item G. Zhu, D. Lizotte, J. Hoey, Scalable approximate policies for Markov decision process models of hospital elective admissions, Artif. Intell. Med. 61 (2014) 21–34. doi:10.1016/j.artmed.2014.04.001.
\item C.B. Browne, E. Powley, D. Whitehouse, S.M. Lucas, P.I. Cowling, P. Rohlfshagen, S. Tavener, D. Perez, S. Samothrakis, S. Colton, A Survey of Monte Carlo Tree Search Methods, IEEE Trans. Comput. Intell. AI Games. 4 (2012) 1–43. doi:10.1109/TCIAIG.2012.2186810.
\item D. Silver, A. Huang, C.J. Maddison, A. Guez, L. Sifre, G. Van Den Driessche, J. Schrittwieser, I. Antonoglou, V. Panneershelvam, M. Lanctot, S. Dieleman, D. Grewe, J. Nham, N. Kalchbrenner, I. Sutskever, T. Lillicrap, M. Leach, K. Kavukcuoglu, T. Graepel, D. Hassabis, Mastering the game of Go with deep neural networks and tree search, Nature. 529 (2016) 484–489. doi:10.1038/nature16961.
\item T. Cazenave, Nested Monte-Carlo expression discovery, in: Proc. 19th Eur. Conf. Artif. Intell., IOS Press, Amsterdam, The Netherlands, 2010: pp. 1057–1058. doi:10.3233/978-1-60750-606-5-1057.
\item G.M.J.B. Chaslot, M.H.M. Winands, H.J. van den Herik, J.W.H.M. Uiterwijk, B. Bouzy, Progressive Strategies for Monte-Carlo Tree Search, New Math. Nat. Comput. 4 (2008) 343–357.
\item M.P.D. Schadd, M.H.M. Winands, M.J.W. Tak, J.W.H.M. Uiterwijk, Single-player Monte-Carlo tree search for SameGame, Knowledge-Based Syst. 34 (2012) 3–11. doi:10.1016/j.knosys.2011.08.008.
\item J.A.M. Nijssen, M.H.M. Winands, Enhancements for multi-player Monte-Carlo tree search, in: Proc. 7th Int. Conf. Comput. Games, Springer-Verlag, Berlin, Heidelberg, 2011: pp. 238–249. doi:10.1007/978-3-642-17928-0$\_$22.
\item M.J.W. Tak, M.H.M. Winands, Y. Björnsson, N-grams and the last-good-reply policy applied in general game playing, IEEE Trans. Comput. Intell. AI Games. 4 (2012) 73–83. doi:10.1109/TCIAIG.2012.2200252.
\item H. Finnsson, Y. Björnsson, Simulation-based approach to general game playing, in: Proc. Twenty-Third AAAI Conf. Artif. Intell., 2008: pp. 259–264.
\item R Core Team, R: A Language and Environment for Statistical Computing, (2018). http://www.r-project.org/.
\item A. Guazzelli, M. Zeller, W. Lin, G. Williams, PMML: An open standard for sharing models, R J. 1 (2009) 60–65. doi:doi=10.1.1.538.9778.
\item T. Fawcett, An introduction to ROC analysis, Pattern Recognit. Lett. 27 (2006) 861–874. doi:10.1016/j.patrec.2005.10.010.
\item S.L. Chen, S.L. Jackson, E.J. Boyko, Diabetes Mellitus and Urinary Tract Infection: Epidemiology, Pathogenesis and Proposed Studies in Animal Models, J. Urol. 182 (2009) 51–56. doi:10.1016/j.juro.2009.07.090.
\item E.W. Dijkstra, A Note on Two Problems in Connexion with Graphs, Numer. Math. 1 (1959) 269–271. doi:10.1007/BF01386390.
\item M.L.G. Fernandez, R.M. Lozano, M.I.G.-Q. Diaz, M.A.G. Jurado, D.M. Hernandez, J.V.B. Montesinos, How Effective Is Orthotic Treatment in Patients with Recurrent Diabetic Foot Ulcers?, J. Am. Podiatr. Med. Assoc. 103 (2013) 281–290. doi:10.7547/1030281.
\item L. Khaodhiar, M. Keane-Ellison, N.E. Tawa, A. Thibault, P.A. Burke, B.R. Bistrian, Iron deficiency anemia in patients receiving home total parenteral nutrition, J. Parenter. Enter. Nutr. 26 (2002) 114–119.
\item F.B. Hu, S. Liu, R.M. van Dam, Diet and risk of Type II diabetes: the role of types of fat and carbohydrate, Diabetologia. 44 (2001) 805–817. doi:10.1007/s001250100547.
\item S.K. Goyal, A. Sharma, Atrial fibrillation in obstructive sleep apnea, World J. Cardiol. 5 (2013) 157–63. doi:10.4330/wjc.v5.i6.157.
\item O. Le Bon, B. Fischler, G. Hoffmann, J.R. Murphy, K. De Meirleir, R. Cluydts, I. Pelc, How significant are primary sleep disorders and sleepiness in the chronic fatigue syndrome?, Sleep Res. Online. 3 (2000) 43–48.
\item M. Fossey, E. Libman, S. Bailes, M. Baltzan, R. Schondorf, R. Armsel, C.S. Fichten, Sleep quality and psychological adjustment in chronic fatigue syndrome, J. Behav. Med. 27 (2004) 581–605. doi:10.1007/s10865-004-0004-y.
\item D. Einhorn, D.A. Stewart, M.K. Erman, N. Gordon, A. Philis-Tsimikas, E. Casal, Prevalence of sleep apnea in a population of adults with type 2 diabetes mellitus., Endocr. Pract. 13 (2007) 355–62. doi:10.4158/EP.13.4.355.
\item E. Tasali, B. Mokhlesi, E. Van Cauter, Obstructive sleep apnea and type 2 diabetes: Interacting epidemics, Chest. 133 (2008) 496–506. doi:10.1378/chest.07-0828.
\item M.T. Naughton, T.D. Bradley, Sleep apnea in congestive heart failure, Clin. Chest Med. 19 (1998) 99–113. doi:10.1016/S0272-5231(05)70435-4.
\item R. Lee, W.T. McNicholas, Obstructive sleep apnea in chronic obstructive pulmonary disease patients, Curr. Opin. Pulm. Med. 17 (2011) 79–83. doi:10.1097/MCP.0b013e32834317bb.
\item M. Shteinberg, D. Weiler-Ravel, Y. Adir, The Overlap Syndrome: Obstructive Sleep Apnea and Chronic Obstructive Pulmonary Disease, Harefuah. 148 (2009) 333–336.
\item T. Kucera, H.H. Shaikh, P. Sponer, Charcot Neuropathic Arthropathy of the Foot: A Literature Review and Single-Center Experience, J. Diabetes Res. 2016 (2016) 1–10. doi:10.1155/2016/3207043.
\item H. Baier, M.H.M. Winands, Nested Monte-Carlo Tree Search for online planning in large MDPs, in: Proc. 20th Eur. Conf. Artif. Intell., 2012: pp. 109–114. doi:10.3233/978-1-61499-098-7-109.
\end{enumerate}

%\bibliography{references}
\end{document}